\documentclass{article}
\usepackage{arxiv}

\usepackage{microtype}
\usepackage{graphicx}
\usepackage{subfigure}
\usepackage{booktabs} % for professional tables
\usepackage[colorlinks=true, allcolors=black, pdftitle={GEEX}]{hyperref}

\usepackage{amsmath}
\usepackage{amssymb}
\usepackage{amsfonts}
\usepackage{amsthm}
\usepackage{aligned-overset}
\usepackage{circledsteps}
\usepackage{multirow}
\usepackage{makecell}

\usepackage[capitalize,noabbrev]{cleveref}

\theoremstyle{plain}
\newtheorem{theorem}{Theorem}

\theoremstyle{definition}
\newtheorem{definition}[theorem]{Definition}

\theoremstyle{remark}

\usepackage[textsize=tiny]{todonotes}

\newcommand\methodAbr{GEEX}
\newcommand\mnist{MNIST}
\newcommand\imagenet{ImageNet}
\newcommand\fm{Fashion-MNIST}
\newcommand\numberthis{\addtocounter{equation}{1}\tag{\theequation}}

\newcommand{\subalign}[1]{%
  \vcenter{%
    \Let@ \restore@math@cr \default@tag
    \baselineskip\fontdimen10 \scriptfont\tw@
    \advance\baselineskip\fontdimen12 \scriptfont\tw@
    \lineskip\thr@@\fontdimen8 \scriptfont\thr@@
    \lineskiplimit\lineskip
    \ialign{\hfil$\m@th\scriptstyle##$&$\m@th\scriptstyle{}##$\hfil\crcr
      #1\crcr
    }%
  }%
}

\renewcommand{\shorttitle}{On Gradient-like Explanation under a Black-box Setting}
\begin{document}

\title{\Large On Gradient-like Explanation under a Black-box Setting: \\
When Black-box Explanations Become as Good as White-box}

% \author{
% \IEEEauthorblockN{Yi Cai}
% \IEEEauthorblockA{\textit{Dept. of Math. and Comp. Science} \\
% \textit{Freie Universität Berlin}\\
% Berlin, Germany \\
% yi.cai@fu-berlin.de}
% \and
% \IEEEauthorblockN{Gerhard Wunder}
% \IEEEauthorblockA{\textit{Dept. of Math. and Comp. Science} \\
% \textit{Freie Universität Berlin}\\
% Berlin, Germany \\
% gerhard.wunder@fu-berlin.de}
% }    
\author{Yi~Cai\\
Dept. of Math. and Comp. Science\\
\textit{Freie Universität Berlin}\\
Berlin, Germany \\
yi.cai@fu-berlin.de \\
\And
Gerhard~Wunder \\
Dept. of Math. and Comp. Science \\
\textit{Freie Universität Berlin}\\
Berlin, Germany \\
gerhard.wunder@fu-berlin.de
}
\maketitle



\begin{abstract}
Attribution methods shed light on the explainability of data-driven approaches such as deep learning models by uncovering the most influential features in a to-be-explained decision.
While determining feature attributions via gradients delivers promising results, the internal access required for acquiring gradients can be impractical under safety concerns, thus limiting the applicability of gradient-based approaches.
In response to such limited flexibility, this paper presents \methodAbr~(gradient-estimation-based explanation), a method that produces gradient-like explanations through only query-level access.
The proposed approach holds a set of fundamental properties for attribution methods, which are mathematically rigorously proved, ensuring the quality of its explanations. 
In addition to the theoretical analysis, with a focus on image data, the experimental results empirically demonstrate the superiority of the proposed method over state-of-the-art black-box methods and its competitive performance compared to methods with full access.
\end{abstract}

% =================== Introduction ===================
\section{Introduction} \label{sec:intro}
Explainability is an increasingly important research topic due to the breakthroughs led by rapidly developing deep learning.
Owing to the growth of hardware computational powers, deep learning models with growing capacities are able to handle tasks in real-world scenarios, which even outperform human experts in certain domains.
% raise applications of AI models in real-world scenarios.
As data-driven models, deep learning solutions at the current stage are distinguished from traditional approaches based on expert systems~\cite{russell2010artificial}.
Data-driven approaches learn their decision rules implicitly from some given data distribution, which conceals decision reasoning from humans.
The shortage of knowledge about the decision-making process puts the deployment of AI models at risk.
A typical example of machine failures is the Clever-Hans-Effect~\cite{johnson1911clever} exposed by previous research~\cite{lapuschkin2019unmasking,geirhos2020shortcut}.
In the context of machine learning, the effect refers to the case that data-driven models learn to use irrelevant features as shortcuts for classification due to an imbalanced data distribution (e.g., watermarks only contained in certain classes of instances because of different data sources).
Models suffering from the Clever-Hans-Effect may perform well in laboratories, but their outcomes can be misleading and totally unreliable in practice.
Apart from the unintentional failure, it has been shown that data-driven models are fragile under adversarial attacks.
These attacks can steer model outcomes by adding artifacts not recognizable by bare human eyes to the targeted input, which makes counteracting adversaries a challenging task.
Given the potential risks, employing these black boxes in crucial application scenarios, such as medical image classification and autonomous driving, can cause unpredictable consequences as they may fail accidentally or intentionally.
% --> For debugging purpose, interested in why a model fails in certain case.
Explainability, a key to the mysterious box of AI and a potential shield against adversarial attacks~\cite{fidel2020explainability,watson2021attack}, is particularly interested in how models make up their minds.

One way of delivering explanatory information is the feature attribution method, which is the focus of this paper. 
The goal of such attribution methods is to determine the contributions of input features to an inquired model outcome, and thus uncover the observations that support its decision.
Existing attribution methods can be categorized as either white-box or black-box approaches depending on their assumption about model accessibility.
As the name implies, white-box explanation methods assume full access to the target model.
Given more details about the inference procedure, they produce precise explanations by investigating the gradient/information flow throughout the target model.
In practice, however, there is no guarantee of detailed internal access to models due to safety and security concerns, which limits the applicability of white-box approaches in real-world scenarios.
Flexibility is another concern.
Modifications are needed when a white-box approach is applied to explain other models that its original design does not consider.
One should not expect a gradient-based approach examining neural networks with backward propagation to uncover the inference process of a tree-based model without adjustments.
Contrary to the full access assumption, black-box explainers require only query-level access, meaning that a to-be-explained model can only be accessed via its input and output interfaces.
As direct investigation into inference procedures becomes unfeasible under a black-box setting, methods of this kind raise queries and explain model decisions indirectly by analyzing the correlation between input features and model outputs.
% details about model structure are not considered by black-box explainers
The loosened accessibility assumption, coupled with the less specific explanation procedure (no prior knowledge about model structure considered), improves the applicability of black-box explainers.
On the other hand, the restricted access poses a challenge in deriving precise explanations, especially when dealing with models handling high-dimensional inputs, such as images.

% \begin{figure}
%     \centering
%     \includegraphics[width=0.5\textwidth]{fig/brief_example.pdf}
%     \caption{Comparison of explanations derived by \textsc{SmoothGrad} and \methodAbr. The first column shows the original inputs and the corresponding predictions. The second and third columns present the explanations from the two different methods.}
%     \label{fig:brief_example}
% \end{figure}

% TODO*: hiden link https://github.com/caiy0220/GEEX
Aiming at combining the strengths of both categories, this paper presents \underline{G}radient-\underline{E}stimation-based \underline{EX}planation (\methodAbr)\footnote{Full code for reproducibility can be found at: \url{https://github.com/caiy0220/GEEX}}, an explanation method producing gradient-like explanations under a black-box setting.  % (Fig.~\ref{fig:brief_example})
By employing gradient estimation, \methodAbr~circumvents the necessity for the full access assumption and, in principle, is applicable to arbitrary models.
This positions \methodAbr~as an alternative for explainability under circumstances where internal details about the target are inaccessible.
In comparison to other black-box explainers, the proposed method produces fine-grained and precise feature attributions rather than fuzzy hot regions. 
The qualitative analysis in the experiment demonstrates that the resulting explanations capture homologous structures when compared to explanations derived from actual gradients.
Most importantly, we theoretically show that \methodAbr~fulfills a set of fundamental properties of attribution methods, ensuring the usefulness and meaningfulness of the resultant explanations.

% The rest of the paper is organized as follows. 
% In Section~\ref{sec:rw}, we discuss state-of-the-art explanation methods covering both categories of model-agnostic and model-specific. 
% Section~\ref{sec:method} elaborate the details of the proposed method \methodAbr.
% To evaluate the explanation performance, we conduct detailed experiments under various test settings in Section~\ref{sec:ex_main}.
% And lastly, Section~\ref{sec:concl} closes the work with a conclusion and discussions on future work.

% =================== Related work ===================
\section{Related work} \label{sec:rw}
Black-box explanation methods are widely adopted in practice due to their flexibility and applicability.
They treat the to-be-explained model as a black-box with its internal functions left out.
The general ideas behind methods of this kind are similar: creating a set of queries by altering the feature values of $x$, then deriving the explanation for the decision $f(x)$ through analysis of the correlation between changes in the inputs and outputs.
LIME~\cite{ribeiro2016should} is one of the most representative methods from this category.
It generates the queries by randomly switching on and off features in the original input and observes the corresponding predictions from the target model.
Based on the observations, LIME then fits an self-explainabile surrogate model (typically in the form of linear regression), a proxy for extracting explanatory information.
For image data, an additional step conducted by LIME is clustering pixels as superpixels according to the similarity of pixel values and their spatial distances~\cite{vedaldi2008quick}, which reduces the search space to a user-defined size.
Simplifying the search space enables LIME to highlight wider regions where the important features locate.

Apparently, grouping pixels can negatively affect explanation quality.
It has been shown that low-level features such as edges and contours are considered informative to classification problems by current deep learning models~\cite{zeiler2014visualizing}.
Superpixel techniques possess the risk of breaking low-level features into diverse components as they inevitably segment pixels along edges.
Consequently, the explanation method may overlook (part of) the divided features or include irrelevant pixels.
RISE~\cite{petsiuk2018rise} overcomes the challenge with mask resizing, an approach that generates smaller initial masks and upsamples them to the target size through bilinear interpolation.
By doing so, RISE is capable of handling any shape of low-level features without expanding the search space, which significantly improves explanation quality.

Compared to black-box methods, more efforts have been spent on white-box approaches when explaining image classifiers, as they sharpen resultant attribution maps owing to the detailed access to the target.
The most straightforward white-box approach directly adopts vanilla gradients~\cite{simonyan2014deep} as explanations.
It traces the partial derivative of the decision function with respect to the input backward throughout the model.
% considers gradients of classification function with respect to the input features as explanations. 
However, previous work~\cite{smilkov2017smoothgrad} shows that explanations based on vanilla gradients can contain heavy noises.
A potential cause of this observation is the rapid derivative fluctuation at small scales~\cite{balduzzi2017shattered}.
\textsc{SmoothGrad}~\cite{smilkov2017smoothgrad} smooths explanation outcomes by applying a Gaussian kernel to the input and averaging over the acquired gradients.
Such a process has a robust denoising effect that positively correlates to the number of samples used during smoothing.
Apart from being a standalone solution for explainability, one can also plug \textsc{SmoothGrad} into other gradient-based approaches for smoothing purposes.
% Difference made by LRP
% which motivates our work, combining the strength of both model-agnostic and model-specific methods.
% But as previously mentioned, methods falling into this category ...
On the other hand, IG (integrated gradients) derives explanations by integrating derivatives from multiple queries over a path~\cite{sundararajan2017axiomatic}.
% Its sampling process distinguishes itself from the previous.
For any input, IG interpolates between the original instance and the pre-defined baseline (usually a zero matrix) to generate queries.
The interpolation interval determines the number of queries to integrate. 
Alternatively, the group of propagation-based methods~\cite{bach2015pixel,montavon2017explaining} nails the denoising challenge by means of propagation rules, which explicitly utilize model structures and complete the explanation process with one single round-trip.

% =================== Method ===================
\section{Gradient Estimation for Explanation} \label{sec:method}
% TODO2: think about whether we formally define black-box setting in the first place.
% This section elaborates the proposed feature attribution method \methodAbr.
Denoting a model function as $f(\cdot)$ and a target input (the explicand) as $\boldsymbol{x}=(x_1,x_2,...,x_p)$, the goal of attribution methods is to determine a vector $\boldsymbol{\xi}\in\mathbb{R}^p$ that decomposes the total contribution $f(\boldsymbol{x})$ into feature attributions, so that the attribution scores satisfy \textit{Completeness}, i.e. $b + \sum_k \xi_k = f(\boldsymbol{x})$.
The bias $b$ is a scalar representing model activation status given the full absence of input features.
Ideally, the bias will have a zero value for properly defined feature absence.
Here, as in the rest of the paper, bold symbols denote vectors, while plain variables signify scalars.
The target function $f(\cdot)$ produces a scalar indicating the model's confidence in its decision-making process. 
In the context of a multi-class setting, the scalar value can be interpreted as the confidence in determining one class, and the resulting explanation reveals the reasoning of the model in deciding whether an explicand belongs to the specific class.
If the internal access of the model is available, the gradient of $f(\cdot)$ with respect to the input features can be readily acquired, facilitating subsequent processing for the derivation of explanations.
Although such a convenience is unfeasible under a black-box setting, luckily, it is still possible to estimate gradients through queries and observations.

\subsection{Gradient Estimation}
Gradient estimation is a group of algorithms designed to approximate the gradient of a function~\cite{mohamed2020monte}, which is widely utilized in black-box optimization problems~\cite{wierstra2014natural}.
It offers an alternative in situations where the acquisition of exact gradients is impractical or computationally expensive.
% Similar to the black-box setting in explainability, gradient estimation offers an alternative when the computation of exact gradients is expensive or 
Different from the attempt to compute gradients through backward propagation of losses, gradient estimation approximates gradients with a search distribution determined by some parameters of interest.
More specifically, it defines gradients as the direction towards lower expected loss with respect to the analyzing target, which is the input features $\boldsymbol{x}$ in the context of explainability.
% The core idea of NES is to acquire search gradients, which are defined as the sample gradients of expected loss, through a parameterized search distribution $\pi(\cdot|x)$.
% More specifically, instead of minimizing the loss function directly with backpropagation, NES implements gradient descent by the sample gradients of loss expectation.
Denoting the loss function by $\mathcal{L}(\cdot)$ and the set of parameters by $\boldsymbol{x}$, the expected loss over the search distribution is defined as:
\begin{equation*}
    J(\boldsymbol{x}):= \mathbb{E}_{\pi(\boldsymbol{z}|\boldsymbol{x})} [\mathcal{L}(\boldsymbol{z})] = \int\mathcal{L}(\boldsymbol{z}) \pi(\boldsymbol{z}|\boldsymbol{x})~\mathrm{d}\boldsymbol{z}
\end{equation*}
where $\pi(\cdot|\boldsymbol{x})$ indicates the probability density function of the search distribution parameterized by $\boldsymbol{x}$ and $\boldsymbol{z}$ denotes samples drawn from the distribution.
Here, the parameter set is denoted as $\boldsymbol{x}$ to emphasize that the analysis target is input features rather than model parameters considered in common settings, which are not available in this case.
The search gradient can be written as follows through simplification using the log-likelihood trick under the assumption that both the loss and the search distribution are continuously differentiable~\cite{mohamed2020monte}:
\begin{align*}
    \nabla_{\boldsymbol{x}}J(\boldsymbol{x})&=
    % \nabla_{x}\mathbb{E}_{\pi(\boldsymbol{z}|\boldsymbol{x})}[\mathcal{L}(f(z))] &=
    \nabla_{\boldsymbol{x}}\int\mathcal{L}(\boldsymbol{z}) \pi(\boldsymbol{z}|\boldsymbol{x})~\mathrm{d}\boldsymbol{z} \\
    &= \int[\mathcal{L}(\boldsymbol{z}) \nabla_{\boldsymbol{x}}\log\pi(\boldsymbol{z}|\boldsymbol{x})] \pi(\boldsymbol{z}|\boldsymbol{x})~\mathrm{d}\boldsymbol{z}\\
    &= \mathbb{E}_{\pi(\boldsymbol{z}|\boldsymbol{x})}[\mathcal{L}(\boldsymbol{z}) \nabla_{\boldsymbol{x}}\log\pi(\boldsymbol{z}|\boldsymbol{x})]
\end{align*}
% which can then be simplified with the log-likelihood trick:
% \begin{align*}
%     \nabla_{\boldsymbol{x}}J(\boldsymbol{x}) &= 
% \end{align*}
The value of the integral above can be empirically approximated with a Monte Carlo estimator given $n$ samples $\{\boldsymbol{z}^{(1)}, \boldsymbol{z}^{(2)}, ..., \boldsymbol{z}^{(n)}\}\sim\pi(\cdot|\boldsymbol{x})$:
\begin{equation} \label{eq:estimate}
    \boldsymbol{\eta}(\boldsymbol{x}) := \nabla_{\boldsymbol{x}}J(\boldsymbol{x}) \approx \frac{1}{n}
    \sum_{i=1}^{n} \mathcal{L}(\boldsymbol{z}^{(i)}) \nabla_{\boldsymbol{x}}\log\pi(\boldsymbol{z}^{(i)}|\boldsymbol{x})
\end{equation}
Once substituting the loss in Equation~\ref{eq:estimate} with the target function, $\eta$ produces the estimated gradient for $f(\cdot)$.
The estimation demonstrates model sensitivities to input features, which, to a certain extent, uncovers the reasoning behind the target decision.
However, the direct usage of gradient estimation is unsatisfactory as it shares several shortcomings with the actual gradient~\cite{sundararajan2017axiomatic}.
\begin{figure}[t]
    \centering
    \includegraphics[width=0.47\textwidth]{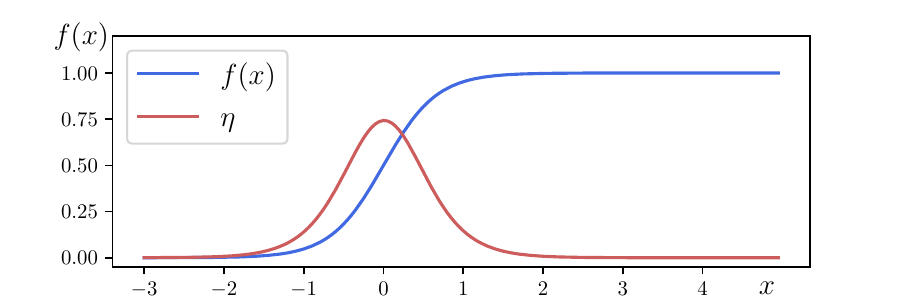}
    \caption{A simple case shows that considering the estimated gradient as an explanation can lead to misleading results.
    Suffering from gradient saturation, the attribution of $x$ converges to 0 as its value increases, conflicting with the truth that the value of the sigmoid function $f(\cdot)$ relies solely on $x$.}
    \label{fig:sensitivity_violate}
\end{figure}
Using estimated gradients as explanations violates \textit{Sensitivity}, a fundamental axiom of attribution methods stating that a feature should receive a non-zero attribution if modifying its value induces a change in the model outcome.
The counterexample in Fig.~\ref{fig:sensitivity_violate} shows that estimated gradients can be trapped by a locally flattened segment of a function, resulting in an underestimation of feature contribution.
Underrating or even overlooking relevant features causes the violation of Sensitivity.
Furthermore, along the $x$-axis of the example, the attribution barely aligns with the total feature contribution represented by $f(x)$. 
The fact that local sensitivities, as indicated by gradients, do not necessarily correlate to actual feature contributions undermines the interpretability of these values in their raw form.

\subsection{Gradient-Estimation-based Explanation} \label{sec:geex}
The failure when employing raw gradient estimation stems from the lack of a reference point that models the absence of features, to which the impact of feature presence can be compared. 
To overcome the aforementioned limitations, we present \methodAbr~(gradient-estimation-based explanation), an attribution method that introduces a baseline and integrates estimations over a straightline path from the baseline to the explicand.
% The baseline values model the absence of features while the original values of the explicand represent feature presence.
Denoting a baseline point as $\boldsymbol{\mathring{x}}$, the straightline path can be written as an interpolation:
\begin{equation*}
    \boldsymbol{x}(\alpha) = \boldsymbol{\mathring{x}} - \alpha(\boldsymbol{x} - \boldsymbol{\mathring{x}})
\end{equation*}
An intuitive implementation of \methodAbr~can be achieved by replacing the actual gradient in IG~\cite{sundararajan2017axiomatic} with the estimation kernel:
\begin{align*}
    \boldsymbol{\xi} := \frac{(\boldsymbol{x} - \boldsymbol{\mathring{x}})}{s} \circ \sum_{j=1}^{s} \boldsymbol{\eta}(\boldsymbol{x}(\frac{j}{s}))
\end{align*}
Replacing the loss with $f(\cdot)$ and expanding $\boldsymbol{\eta}$ yield:
\begin{equation} \label{eq:rawgeex}
    \boldsymbol{\xi} = \frac{(\boldsymbol{x} - \boldsymbol{\mathring{x}})}{n\cdot s} \circ \sum_{j=1}^{s} \sum^{n}_{i=1} f(\boldsymbol{z}^{(i)}) \nabla_{\boldsymbol{x}}\log\pi(\boldsymbol{z}^{(i)}|\boldsymbol{x}(\frac{j}{s}))
\end{equation}
Although the choice of the search distribution is left unspecified here, we restrict $\boldsymbol{x}$ to be the location parameter, and the search distribution is required to have a mean at its location parameter, which is a necessity for unbiased estimation (see Appendix~\ref{sup:completeness} for theoretical details).
% Although the choice of the search distribution is not yet specified, it should be noted that the distribution should have a mean at $\boldsymbol{x}$ for unbiased estimation (see Appendix~\ref{sup:completeness} for more information).
% There are other possibilities~\cite{TODO*} that can reduce the variance of the estimator $\boldsymbol{\eta}$, but we defer the discussion to future work.
A surprising fact is that \methodAbr~inhabits more or less all of the properties of IG.
In fact, the resultant method complies with a set of four fundamental axioms as stated in Theorem~\ref{theorem:4axioms} (the proof and further details of Theorem~\ref{theorem:4axioms} are located in Appendix~\ref{sup:axioms}).
Please note that the fulfillment of the axioms comes true when enough samples have been drawn, so the statement should be interpreted in a probability sense.

\begin{theorem} \label{theorem:4axioms}
\methodAbr, a path method built upon estimated gradients, satisfies Sensitivity, Insensitivity, Implementation Invariance, and Linearity.
\end{theorem}

As the significance of \textit{Sensitivity} has been shown before, the three remaining axioms also hold practical meanings to the proposed method. 
\textit{Insensitivity} (called \textit{Dummy} in~\cite{friedman2004paths}) is a property that measures attributions to features having no impact. 
Opposite to Sensitivity, a violation of Insensitivity results in an overestimation of feature importance; failure to fulfill either of the two results in misleading explanation outcomes.
Meanwhile, \textit{Implementation Invariance}, a key concept especially for black-box explainers, ensures the applicability of \methodAbr.
\textit{Linearity} appears trivial among the four axioms as it does not directly relate to explanation quality.
However, Linearity enables the decomposition of non-interacting features (features having no interaction with each other in the target function).
Such a decomposition divides a high-dimensional feature space into subspaces with lower dimensionalities.
For a function consisting of $m$ terms, the variance of the gradient estimator deployed by \methodAbr~is of the order $O(m^2)$~\cite{mohamed2020monte}.
Feature space decomposition that linearly reduces the number of terms results in a quadratic reduction of estimation variance.
Although developing a detailed decomposition strategy exceeds the scope of this paper, we argue that the fulfillment of Linearity holds the potential to enhance estimation precision and computational efficiency, which indirectly contribute to explanation quality.

Having covered various axioms, of particular note is the \textit{Completeness} of \methodAbr~stated in Theorem~\ref{theorem:completeness}.
Being an approximation of the path integral, the sum of feature attributions determined by \methodAbr~converges in probability to the prediction difference between the baseline and the explicand, i.e. $f(\boldsymbol{x})-f(\boldsymbol{\mathring{x}})$.
Viewing the prediction at $\boldsymbol{\mathring{x}}$ as the bias $b$, the Completeness will become evident with sufficient observations (for the detailed proof, see Appendix~\ref{sup:completeness}).
Satisfying Completeness is fundamental, although many do not, for attribution methods.
This property upholds the practical meaning of attribution score -- a value indicating the proportion of feature contribution to model outcome.

\begin{theorem} \label{theorem:completeness}
(Completeness)
The explanation derived by \methodAbr~is complete regarding the model outcome $f(\boldsymbol{x})$.
% The sum of feature attributions determined by \methodAbr~converges in probability to the total contribution defined by the expected model outcome $J(\boldsymbol{x})$.
\end{theorem}

Again, given the explanation by \methodAbr~as an empirical approximation, the statement should be interpreted with a probability perspective.
It is noteworthy that the approximation error has two sources: the error associated with the line integral approximation and the error of the gradient estimator.
Although Completeness is desired, the iterative estimation process poses a challenge in deploying \methodAbr, as both sources contribute to the overall error of the explainer.
Under limited computational resources, managing the allocation of efforts between the two estimators -- the interpolation steps $s$ and the sample set size $n$ -- for optimizing the ultimate precision of the explainer can be demanding.
% In the upcoming section, we propose and discuss a solution that mitigates the practical difficulty.
Moreover, even with the puzzle of hyperparameter selection solved, the performance of the proposed approach is bounded by IG from above.
To address the challenges, the next section introduces the way of improving the practical usefulness of \methodAbr, enabling it to go for beyond IG. 

\subsection{Noise Sampling and Computational Efficiency} \label{subsec:noise}
% TODO2: mentioned somewhere that the interpolation is upper bounded by IG, while the smoothed is not!!

% Before delving into the improvement of the explainer, 
First, we clarify the role of $\boldsymbol{x}$ in the search distribution.
In principle, $\boldsymbol{x}$ can represent any parameter of any distribution, on condition that $\pi(\cdot|\boldsymbol{x})$ is continuously differentiable in $\boldsymbol{x}$.
However, in addition to ensuring unbiased estimation as stated in the last section, considering $\boldsymbol{x}$ as the location parameter of the search distribution brings practical convenience to the explanation process.
This is particularly advantageous when handling different explicands, as it allows the usage of an pre-generated mask set during the explanation procedure.
% However, during the explanation procedure, the estimator $\boldsymbol{\eta}$ can assume significantly different values for $\boldsymbol{x}$ when handling distinct explicands.
% To facilitate sample set construction, thereby promoting computational efficiency, we constrain $\boldsymbol{x}$ to be the location parameter of a distribution.

For a standard distribution $\pi_{\boldsymbol{\theta}}(\boldsymbol{z}|\boldsymbol{0})$, the search distribution for a concrete explicand $\boldsymbol{x}$ from the location family holds $\pi_{\boldsymbol{\theta}}(\boldsymbol{z}|\boldsymbol{x}) = \pi_{\boldsymbol{\theta}}(\boldsymbol{z}-\boldsymbol{x}|\boldsymbol{0})$.
Here, $\boldsymbol{\theta}$ represents the remaining parameters, distinct from $\boldsymbol{x}$, that describe the distribution.
The parameter set $\boldsymbol{\theta}$ can be a hyperparameter of the explainer, but we omit it for simplicity as it is irrelevant to the following discussion.
Designating $\boldsymbol{x}$ as the location parameter allows a pre-construction of the sample set and pre-computation of the log derivative with the standard distribution according to the revision of Equation~\ref{eq:rawgeex}:
\begin{align*}
    \boldsymbol{\xi} &= \frac{(\boldsymbol{x} - \boldsymbol{\mathring{x}})}{n\cdot s} \circ \sum_{j=1}^{s} \sum^{n}_{i=1} f(\boldsymbol{z}^{(i)}) \nabla_{\boldsymbol{x}}\log\pi(\boldsymbol{z}^{(i)}-\boldsymbol{x}(\frac{j}{s})|\boldsymbol{0}) \\
    &= \frac{(\boldsymbol{x} - \boldsymbol{\mathring{x}})}{n\cdot s} \circ \sum_{j=1}^{s} \sum^{n}_{i=1} f(\boldsymbol{x}(\frac{j}{s}) + \boldsymbol{\epsilon}^{(i)}) \nabla_{\boldsymbol{x}}\log\pi(\boldsymbol{\epsilon}^{(i)}|\boldsymbol{0})
\end{align*} 
where $\boldsymbol{\epsilon}=\boldsymbol{z}-\boldsymbol{x}$ denotes the pre-generated mask.
The construction of the mask set $\{\boldsymbol{\epsilon}^{(i)}\}$ is a one-time effort, and it can be applied to arbitrary explicand-baseline pairs.
This is possible because sampling for gradient estimation is decoupled from the concrete value of $\boldsymbol{x}$.
This convenience facilitates the application of more complicated sampling strategies aiming at variance reduction, such as mirror sampling~\cite{brockhoff2010mirrored} considered in this work, or potentially other computationally more expensive ones like orthogonal coupling~\cite{choromanski2019unifying}.
% In this work, we considered mirror sampling, an approach enhancing estimation robustness by adding symmetric pairs $\{\boldsymbol{\epsilon}, -\boldsymbol{\epsilon}\}$ to the mask set.

\begin{figure}[t]
    \centering
    \includegraphics[width=0.47\textwidth]{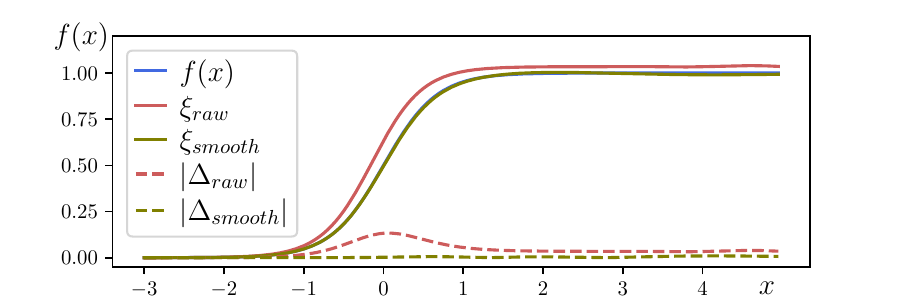}
    \caption{Given a baseline $f(-3)\approx0$, the smoothed version of \methodAbr~better approximates the actual contribution of the input feature with the same amount of observations.
    While the red solid line corresponds to explanations from the interpolation-based \methodAbr, the green line represents the results from the "smoothed" version, almost overlapping the actual contribution depicted by the blue line. The dashed line indicates the error of the derived explanation compared to the ground truth given by the total contribution $f(x)$.}
    % A simple case shows that considering the estimated gradient as an explanation can lead to misleading results.
    % Suffering from gradient saturation, the attribution of $x$ converges to 0 as its value increases, conflicting with the truth that the value of the sigmoid function $f(\cdot)$ relies solely on $x$.
    \label{fig:smoothen}
\end{figure}

\begin{figure*}
    \centering
    % trim={1cm 2cm 1cm 1cm}, 
    \includegraphics[width=.94\textwidth]{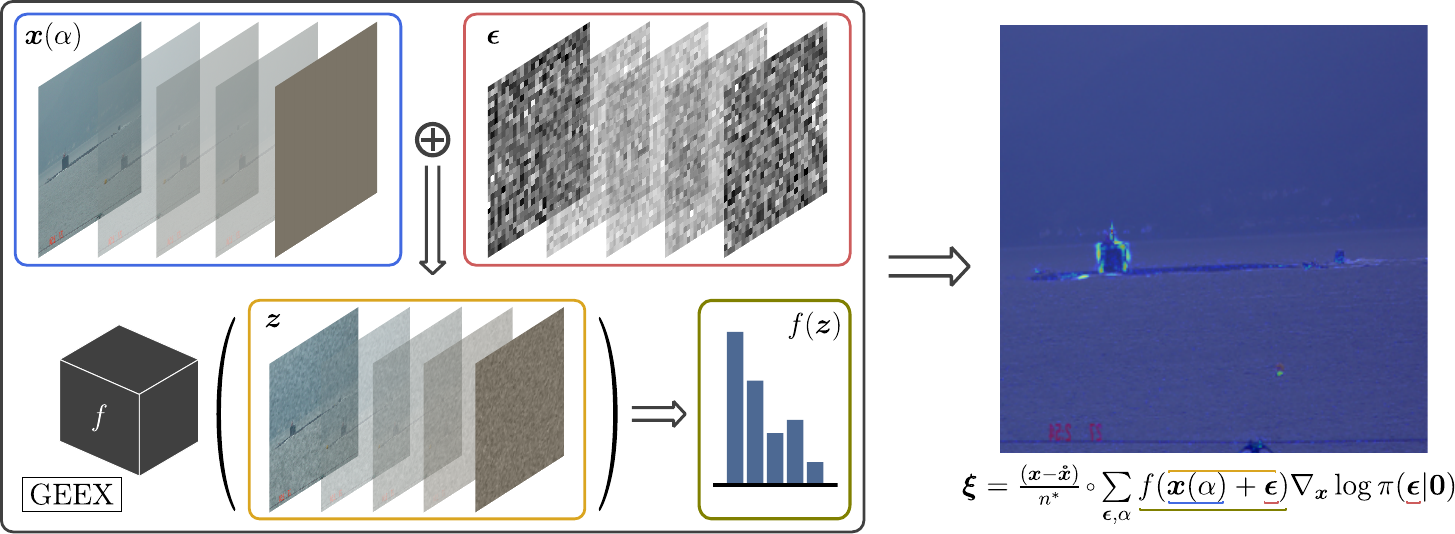}
    \caption{Overview of \methodAbr. 
    A query $\boldsymbol{z}$ is determined by the sampled noise $\boldsymbol{\epsilon}$ and the position $\alpha$ on the path.
    % constructed by applying the upsampled noises $\epsilon$ from the initial mask set $\hat{\epsilon}$ to the input $x$. 
    The final explanation $\boldsymbol{\xi}$ (on the right, overlaid with the original input) is derived through the observations $\{f(\boldsymbol{z})\}$ and the pre-computed log derivatives.}
    \label{fig:geexOverview}
\end{figure*}

% in the two levels of approximation
Moreover, recognizing the decoupling of sampling from the location parameter, the sums from the two levels of approximation can be merged:
\begin{equation*}
    \boldsymbol{\xi} = \frac{(\boldsymbol{x} - \boldsymbol{\mathring{x}})}{n^*} \circ \sum_{
    % \boldsymbol{\epsilon}\sim\pi(\cdot|\boldsymbol{0}), ~\alpha\sim \mathcal{U}_{[0, 1]}
    \substack{\boldsymbol{\epsilon}\sim\pi(\cdot|\boldsymbol{0}) \\ \alpha\sim \mathcal{U}_{[0, 1]}}
    } f(\boldsymbol{x}(\alpha) + \boldsymbol{\epsilon}) \nabla_{\boldsymbol{x}}\log\pi(\boldsymbol{\epsilon}|\boldsymbol{0})
\end{equation*}
where $n^*$ denotes the number of queries generated by $(\boldsymbol{\epsilon}, \alpha)$ pairs.
Although the straightforward rewriting does not alter any underlying concepts, it streamlines hyperparameter selection.
The form can be interpreted as the cumulative sum of dense \textit{one-sample gradient estimators} along the integral path.
More importantly, merging the terms for integral approximation and gradient estimation improves explanation quality by providing a ``smoother'' approximation of the path integral without compromising the precision of gradient estimations.
% Aligning with the discussion, explanations by GEEX outperforms IG when sufficient observations are provided   
Fig.~\ref{fig:smoothen} provides an illustrative example of the smoothed approximation.
The key factor, allowing the improvement, is that observations from estimators $\boldsymbol{\eta}(\boldsymbol{x}(\alpha))$ for neighboring instances on the path share information, which positively contributes to the estimation precision of each other.
The proof of the claim is given in Appendix~\ref{sup:complement_neighbors}.
The nature that \methodAbr~can smoothly approximate the path integral allows it to even surpass white-box explainers under certain circumstances (see further discussions in Section~\ref{sec:exp_aopc}).
Figure~\ref{fig:geexOverview} gives an overview of the explanation procedure.
% TODO2: especially useful when the gradient of the model function is not uniformly distributed? 

Though the construction of a mask set $\{\boldsymbol{\epsilon}^{(i)}\}$ is straightforward in most cases, we suggest mask smoothing for the sampling when dealing with high dimensional inputs, particularly referring to high-resolution images.
Denoting the initial masks as $\boldsymbol{\hat{\epsilon}}$ and a blurring kernel as $\boldsymbol{w}$, the post-processing that finalizes the masks can be described as $\boldsymbol{\epsilon} = \frac{\boldsymbol{w}}{||\boldsymbol{w}||_F} * \boldsymbol{\hat{\epsilon}}$, where $||\boldsymbol{w}||_F$ is the Frobenius norm of the filter that normalizes the amplitude of perturbation, ensuring it is invariant to the convolution operator.
In addition to the denoising effect by mitigating artifacts in adjacent pixels of masks, the filter softly groups spatially close pixels following the prior knowledge that adjacent pixels form low-level features.
By applying similar changes to adjacent pixels simultaneously, soft grouping increases the possibility of removing a local pattern compared to conducting pixel-wise perturbation.
In the case of high-dimensional explicands, such a convenience helps expose model sensitivities to the absence of local patterns, thus facilitating the identification of relevant pixels.
However, it should be noted that the grouping does not stick to the assumption that feature values should be sampled independently for gradient estimation. 
Therefore, the application of mask smoothing raises a trade-off between the usefulness and correctness of resultant explanations and is preferred only when explaining high-dimensional explicands.

% reduce dimensionality of the sampling space 
% mask out a high-level feature --> exploits influential features 
% reducing noises
% --------------------
% The motivation of doing so?
% Trade-off between usefulness and correctness.
% Softly grouping pixels following the prior knowledge that neighboring pixels forming a feature

\section{Experiments} \label{sec:ex_main}
With a focus on image data, we test the proposed method with neural networks trained on three popular and publicly available image datasets.
The experimental environment (including hardware and software settings) is described in Supplement~S-I.

\subsection{Experimental Details}
\textbf{Dataset:} The datasets considered during the experiments are: \mnist~\cite{lecun1998gradient}, \fm~\cite{xiao2017fashion}, and \imagenet~\cite{russakovsky2015imagenet}.
The selection includes two grayscale datasets and one full-colored dataset with a notably larger input size.
For each dataset, we train a classifier with its training set and evaluate explanations for model decisions on the test set.

\textbf{Classifier:} For \mnist~and \fm, a simple CNN is trained.
The model comprises two convolutional layers with a kernel size of 5, concatenated by three dense layers with sizes of 120, 84, and 10, respectively.
The inputs from the two datasets have a shape of $28\times28$.
For \imagenet, Inception V3\footnote{
A pre-trained version from \imagenet~is used without additional training, publicly available at: \url{https://pytorch.org/vision/stable/models/inception.html}
}~\cite{szegedy2016rethinking} is adopted, which takes inputs of size $299\times299$.
Considering various configurations of the to-be-explained system enables the comparison of explainer performances across explanation tasks with different levels of complexity.

\textbf{\methodAbr:} 
A Gaussian distribution serves as the search distribution for \methodAbr, and the number of queries $n^*$ is fixed to $5k$ across all test settings.
The deviation $\sigma$, which determines the spread of the Gaussian, is configured as $1.0$ for \mnist~and \fm~observing the polarized distribution of their pixel values.
For \imagenet, where pixel values are more evenly distributed, $\sigma$ is set to $0.3$.
Regarding the baseline $\mathring{\boldsymbol{x}}$, a zero matrix is employed when explaining decisions on grayscale images, whereas the baseline for \imagenet~is explicand-specific. 
For each explicand from \imagenet, the baseline is the blurred version of itself.
To ensure a fair comparison, these baseline configurations also apply to competitors who involve a baseline during their explanation procedures.
% TODO2: mention mirror sampling here, maybe?
Besides, to mitigate the estimation noises caused by feature space expansion, mask smoothing is implemented through a Gaussian filter with a kernel size of $5$ and a deviation of $0.7$ when tested on \imagenet.
Due to the space limitation, further details regarding hyperparameter selection are reported in Supplement S-III.
% The effects of the hyperparameter selection are reported in Supplement~\ref{sup:effect}.

\textbf{Competitor:} 
We consider the gradient estimator (noted as GE) as a competitor by directly interpreting gradient estimations as explanations. 
Its comparison to \methodAbr~illustrates the importance of fulfilling the named properties.
The remaining competitors, including two white-box approaches and two black-box explainers, are listed below.
For all black-box competitors, the number of queries is identical to the setting for \methodAbr.
\begin{itemize}
    \item SG (\textsc{SmoothGrad})~\cite{smilkov2017smoothgrad}: a white-box approach interprets raw gradients as explanations.
    % Besides, we consider it the \textbf{baseline} due to its direct connection to gradients.
    \item IG~\cite{sundararajan2017axiomatic}: a white-box approach integrates gradients of entries over a straightline path from the explicand to a baseline.
    \item LIME~\cite{ribeiro2016should}: a black-box approach explains a target model through a linear surrogate trained to mimic the observed behaviors of the target.
    % trains a linear surrogate to mimic the local classification behavior of the target model for deriving explanations.
    \item RISE~\cite{petsiuk2018rise}: a black-box explainer determines feature attribution according to the expected impacts of input features on prediction outcomes.
\end{itemize}

\subsection{Comparison to White-box Explanation}
We commence with the evaluation of the proposed method with a qualitative assessment of the derived explanations.
Sample explanations from various test settings are listed in Fig.~\ref{fig:qualitative_example} (additional examples can be found in Supplement~II), where each column is an example set.
% demonstrate the proximity between explanations from \methodAbr~and IG 
The first image in an example set presents the explicand with the model prediction at the top, followed by the corresponding explanations from three chosen explainers visualized through saliency map.
The explainers comprise IG, \methodAbr, and RISE.
The inclusion of IG and RISE serves the purpose of showcasing \methodAbr's capability in producing gradient-like explanations under a black-box setting and emphasizing its improvements compared to a state-of-the-art black-box explainer.

\begin{figure*}[ht!]
    \centering    
    \includegraphics[width=.99\textwidth]{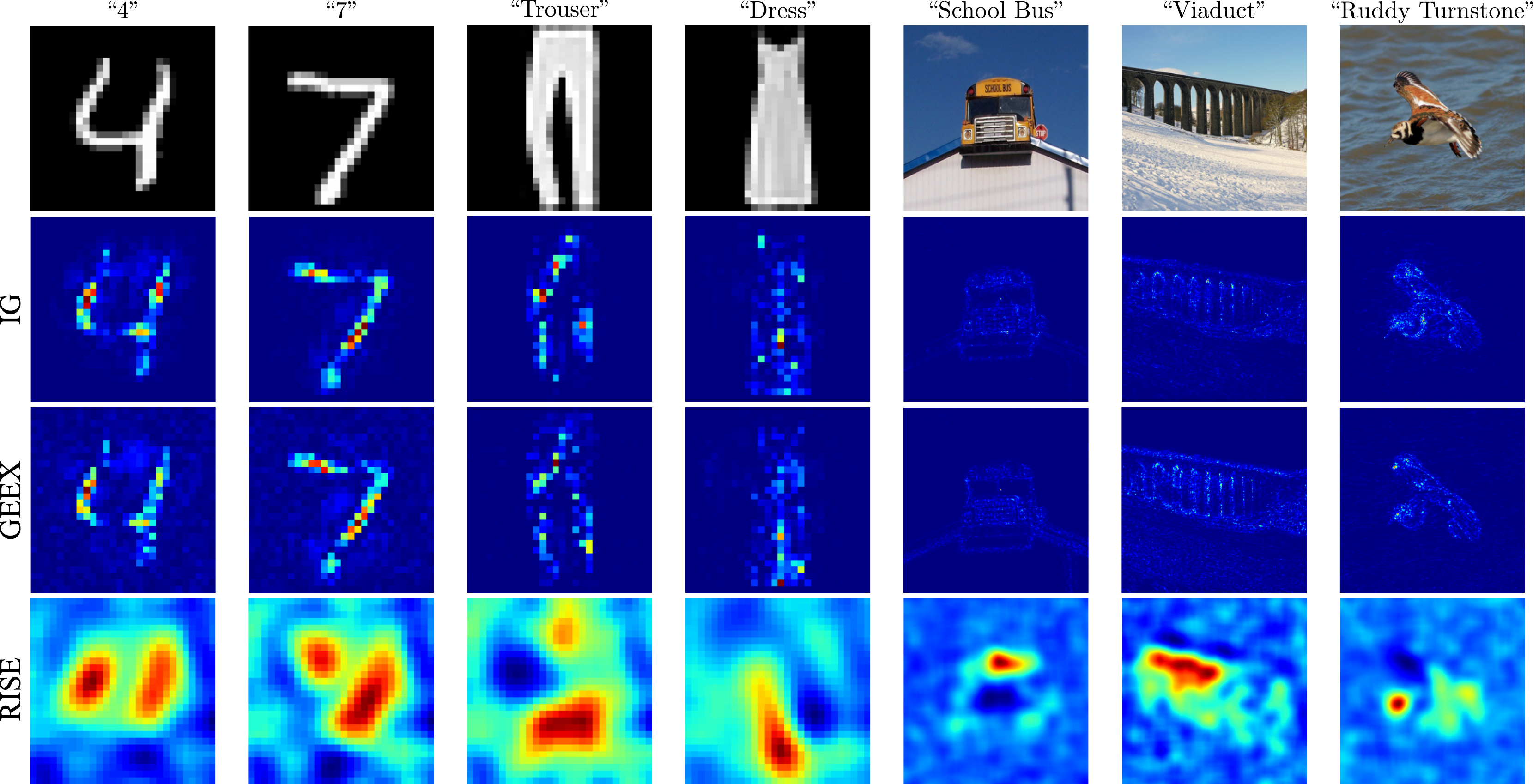}
    \caption{Sample explanations from the selected competitors}
    \label{fig:qualitative_example}
\end{figure*}

The provided examples contain explanations for four explicands with a smaller input size and three entries from a significantly higher-dimensional feature space.
Across all sample sets, proximity between explanations from \methodAbr~and IG is observed.
Specifically, both \methodAbr~and IG attribute the model's decision to the presence of pixels on the pant legs for the third explicand, predicted as ``trouser'', while pinpointing the bottom-middle located pixels as the most influential for predicting a ``dress'' in the fourth example. 
The insights garnered from the two explainers suggest that the classifier trained on \fm~considers the presence of a gap in the bottom-middle region between two bright areas as a distinguishing feature between a trouser and a dress.
However, the explanations from RISE fall short of reaching such a conclusion.
Similar observations can be found in the sample explanations for InceptionV3.
While the outcomes from RISE highlight fuzzy hot regions with noticeable background noises, the delivery of the proposed approach demonstrates fine-grained feature structures that steer the model's predictions.
These include details such as the texture of the intake grille for the classified ``school bus'', the contour of the ``viaduct'', and the highlighted face as well as wing areas of the ``ruddy turnstone''.
Again, the explanations by \methodAbr~capture homologous attribution structures to the results from IG when dealing with more challenging tasks, consistent with the previously discussed performance in simpler test cases.

The qualitative examples intuitively show the performance of the proposed method.
Nevertheless, human assessments for explanation evaluation are unscalable.
Exhaustively evaluating explanation approaches with human efforts can be more than just expensive in practice.
In fact, doing so can introduce human-sourced biases into the evaluation process as individuals may interpret explanations differently.
For instance, given a model suffering from the Clever-Hans effect, which occasionally uses irrelevant features for prediction, an explainer correctly locating such mistakes can be underrated during human assessments due to mismatches between human domain knowledge and the actual model behaviors.

\subsection{Quantitative Evaluation for Effectiveness}  \label{sec:exp_aopc}
\begin{table*}[tbp]
\caption{The normalized AOPC scores by evaluation via deletion, higher is better.}
    \centering
    \begin{tabular}{cc||cc|cccc|c}
        \hline & & & & & & & & \\ [-3mm]
        Classifier & Replacement & SG &  IG & RISE & LIME & GE & \methodAbr & Random \\[0.5mm]
        \hline & & & & & & & \\ [-3mm]
        \multirow{2}{*}{\makecell[c]{CNN \\ (\mnist)}} & Baseline & 0.8838 & 0.9434 & 0.9101 & 0.8653 & 0.8833 & \underline{\textbf{0.9466}} & 0.3085 \\[0.5mm]
        & Gaussian & 0.8452 & 0.9415 & 0.8896 & 0.8692 & 0.8519 & \underline{\textbf{0.9486}} & 0.3695 \\[0.5mm]
        % 0.8913 & \textbf{0.9347} & 0.8596 & 0.8696 & 0.7928 & \underline{0.9338} & 0.3658 \\[0.5mm]
        \hline & & & & & & & \\ [-3mm]
        \multirow{2}{*}{\makecell[c]{CNN \\ (F-MNIST)}} & Baseline & 0.8399 & 0.9275 & 0.8931 & 0.8167 & 0.8379 & \underline{\textbf{0.9350}} & 0.3567 \\[0.5mm]
        & Gaussian & 0.8341 & 0.9219 & 0.8708 & 0.8085 & 0.8341 & \underline{\textbf{0.9362}} & 0.4293 \\[0.5mm]
        \hline & & & & & & & \\ [-3mm]
        \multirow{2}{*}{\makecell[c]{InceptionV3 \\ (\imagenet)}} & Baseline & 0.3781 & \textbf{0.8805} & 0.7659 & 0.6928 & 0.3806 & \underline{0.7952} & 0.4003 \\[0.5mm]
        & Gaussian & 0.8557 & \textbf{0.9155} & 0.8699 & 0.8837 & 0.7289 & \underline{0.9058} & 0.7434 \\[0.5mm]
        % 0.8766
        \hline
        \multicolumn{9}{l}{\footnotesize{$^{\mathrm{*}}$The overall best performances are in \textbf{bold} and the highest scores among black-box explainers are \underline{underlined}.}}
    \end{tabular}
    \label{tab:aopc}
\end{table*}

To quantify the performance of explainers objectively, this section evaluates explanation quality via a widely adopted scheme -- evaluation via deletion~\cite{samek2016evaluating}.
% Compromising to the absence of a ground truth in the context of explainability, 
The evaluation process follows an intuitive yet effective idea: the removal of relevant features should induce larger drops in prediction confidence.
More specifically, evaluation via deletion removes pixels sequentially in descending order according to their attribution scores.
The changing trend of prediction confidence draws a curve throughout the deletion process, and the area over perturbation curve (AOPC) is considered as a metric to quantify the effectiveness of an explanation.
To make the metric independent from the scale of model outcomes, the normalized AOPC is computed, i.e. the cumulative sum of dropping ratios: $AOPC=\frac{1}{l}\sum^l_{i=1}(1 - \frac{f(x^{(i)})}{f(x)})$, where $x^{(i)}$ denotes a variant of $x$ with its top-$i$ pixels masked out.
% The AOPC score reflects not only the ability of an explainer to identify relevant features but also the ranking correctness of their contributions.
One last thing to be clarified is the deletion operation.
Following the discussion in Section~\ref{sec:geex}, replacing a feature value with a corresponding baseline value is a natural way of defining the deletion.
As some competitors do not actively use the chosen baselines, for a fair comparison, we also sample the replacement value from Gaussian as an alternative defining the deletion.

Table~\ref{tab:aopc} reports the AOPC scores of competitors in all test settings, considering both definitions of the replacement value.
The table groups explainers according to their accessibility assumptions.
Alongside the named competitors, the AOPC scores of removing pixels in purely random order are also reported as a reference, illustrating the effectiveness of the derived explanations.
Ideally, explanations delivering useful information are expected to achieve higher scores than random deletion.
However, this is not the case for SG and GE in explaining decisions from InceptionV3.
Their explanations, directly using either actual or estimated gradients, suffer from gradient saturation, leading to the overlooking of relevant features and subsequently limited performance.
On the contrary, the fulfillment of Completeness and Sensitivity results in the competitive performance of \methodAbr.
According to the AOPC scores, the proposed method consistently surpasses other black-box explainers across all test settings.
The higher scores indicate that the assigned feature attributions correctly reflect to their actual contributions.
% its explanations identify relevant features with the assign attributions correctly correlate to their actual contribution.

\begin{figure}[t]
    \centering
    \includegraphics[width=0.47\textwidth]{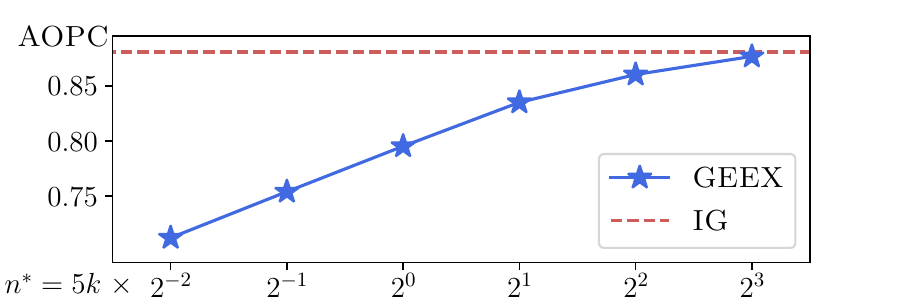}
    \caption{For InceptionV3, \methodAbr~achieves an AOPC score converging to IG when $n^*$ increases.}
    \label{fig:convergence}
\end{figure}

Compared to IG, \methodAbr~achieves similar scores, aligning with the observations of their visually similar saliency maps in the qualitative assessment.
For the simpler test cases, our approach even achieves better performances, which should be interpreted as an improvement brought by the smoother approximation of the path integral.
While the white-box explainer approximate the path integral with sparse interpolation relying on accurate but expensive gradients (depending on model complexity), \methodAbr~achieves a superior approximation with dense one-sample estimators that are computationally more efficient.
Regarding the results on \imagenet, the larger feature space poses a challenge to all black-box approaches.
As a result of higher gradient estimator variance caused by feature space expansion, \methodAbr~falls behind IG.
In this case, more observations are required to maintain the same level of estimation precision.
Figure~\ref{fig:convergence} illustrates the convergence of \methodAbr's performance towards IG as the number of queries increases.

% As the experiment delivers promising results, it is noteworthy that computational complexity can be a potential concern of GEEX, which is similar to other BBs.
% quadratic growth...
% although computation of GEEX can be highly parallelized and distributed to several machine to meet real-time constrain, we believe using Linearity that decomposes feature space is the ultimate solution that solves the puzzle.
% ---> In conclusion: Talk about efficiency as future work.
    % ---> possible application in other data forms

% =================== Conclusion ===================
\section{Conclusion} \label{sec:concl}
% Also, we show that the estimated gradients can be substituted into the integrated gradients, ..., although the results have a strong dependency on the estimation quality.
In this work, we propose \methodAbr, an approach deriving gradient-like explanations under a black-box setting.
It fulfills a set of fundamental properties of attribution methods, including \textit{Completeness} and \textit{Sensitivity}, thereby settling a theoretical guarantee for explanation quality.
Alongside the theoretical analysis, the experimental results on three public datasets empirically show the competitive performance of the proposed method.
In addition to surpassing all competitors in the simple test cases, the performance of \methodAbr~also converges to the best score achieved in a white-box setting when acquiring sufficient observations.
Although the computational expense can be a concern as for other black-box approaches, the computations in \methodAbr~are highly parallelized, which allows it to meet any potential real-time requirement by distributing the workloads to distributed agents.
Moreover, we believe that reducing computational complexity through feature space decomposition, guided by the \textit{Linearity} property, addresses the last piece of the puzzle and should be considered a direction of future works.

% The resultant approach ... properties ... theoretical guarantee.
% In addition to the theoretic analysis, the design experiments empirically demonstrate the promising performance of the proposed method.
% Although the results are promising, it is noteworthy that while gradient-based approaches have a complexity linearly correlates to the number of learnable parameters, the sample set size for \methodAbr~is in at least quadratic (or more rapid depending on $f(\cdot)$) growth for the same explanation quality with feature space expansion, which can pose practical difficulty.
% linearity as a potential ... solving the puzzle should be given more thoughts as future work.

% Although the experiments focus on image data because of the interest in deriving black-box explanations in high-dimensional feature spaces, we want to note that the general idea is applicable to other data forms.

\section*{Acknowledgement}
Yi~Cai and Gerhard~Wunder were supported by the Federal Ministry of Education and Research of Germany (BMBF) in the program of “Souverän. Digital. Vernetzt.”, joint project “AIgenCY: Chances and Risks of Generative AI in Cybersecurity”, project identification number 16KIS2013. Gerhard~Wunder was also supported by BMBF joint project “6G-RIC: 6G Research and Innovation Cluster”, project identification number 16KISK020K.

\bibliographystyle{apalike}
\bibliography{example_paper}
% \include{example_paper}

%%%%%%%%%%%%%%%%%%%%%%%%%%%%%%%%%%%%%%%%%%%%%%%%%%%%%%%%%%%%%%%%%%%%%%%%%%%%%%%
%%%%%%%%%%%%%%%%%%%%%%%%%%%%%%%%%%%%%%%%%%%%%%%%%%%%%%%%%%%%%%%%%%%%%%%%%%%%%%%
% APPENDIX
%%%%%%%%%%%%%%%%%%%%%%%%%%%%%%%%%%%%%%%%%%%%%%%%%%%%%%%%%%%%%%%%%%%%%%%%%%%%%%%
%%%%%%%%%%%%%%%%%%%%%%%%%%%%%%%%%%%%%%%%%%%%%%%%%%%%%%%%%%%%%%%%%%%%%%%%%%%%%%%
\newpage
\onecolumn

\appendix

\section{Proof of Completeness} \label{sup:completeness}
Differing from the order of introducing the properties as presented in Section~\ref{sec:method}, we first show the \textit{Completeness} of \methodAbr, which is stated in Theorem~\ref{theorem:completeness}.
Proving Completeness brings significant convenience to the proof of the remaining axioms.
% This section gives the proof for the Completeness of \methodAbr, which is stated in Theorem~\ref{theorem:completeness}.
% Stated by Theorem~\ref{theorem:completeness}, 
Defining the full absence with the baseline, the Completeness regarding the prediction outcome $f(\boldsymbol{\dot{x}})$ can be proved by showing:
\begin{equation*}
    f(\boldsymbol{\mathring{x}}) + \sum_{k=1}^{p} \xi_k = f(\boldsymbol{\dot{x}})
\end{equation*} 
where $\boldsymbol{\dot{x}}$ denotes the explicand to avoid the potential confusion between the variable $\boldsymbol{x}$ and its concrete value given by $\boldsymbol{\dot{x}}$.
The premise for \methodAbr~satisfying Completeness is that the search distribution for the gradient estimator has a mean at its location parameter determined by the explicand:
\begin{equation} \label{eq:unbias_distribution}
    \mathbb{E}_{\pi(\boldsymbol{z}|\boldsymbol{x})}[\boldsymbol{z}] = \boldsymbol{x}
\end{equation}
Through the proof of Completeness, we also show that an unbiased search distribution is a necessity for an unbiased gradient estimation.

\begin{proof}
    In \methodAbr, the attribution of the $l$-th feature is given by:
    \begin{align*}
        \xi_l &= \frac{(\dot{x}_l - \mathring{x}_l)}{s} \cdot \sum_{j=1}^{s} \eta_l(\boldsymbol{\dot{x}}(\frac{j}{s})) \\
        % &= \frac{(\dot{x}_l - \mathring{x}_l)}{s} \cdot \sum_{j=1}^{s} \frac{1}{n} \sum^{n}_{i=1} f(\boldsymbol{z}^{(i)}) \frac{\partial}{\partial x_l} \log\pi(z_l^{(i)}|\dot{x}_l(\frac{j}{s})) \\
        &= \frac{(\dot{x}_l - \mathring{x}_l)}{s} \cdot \sum_{j=1}^{s} \frac{1}{n} \sum^{n}_{i=1} f(\boldsymbol{z}^{(i)}) \frac{\partial \log\pi(z^{(i)}_l|x_l)}{\partial x_l}\Big{|}_{x_l=\dot{x}_l(\frac{j}{s})} \\
        \overset{\boldsymbol{z}=\boldsymbol{\dot{x}}(\frac{j}{s})+\boldsymbol{\epsilon}}&{=} \frac{(\dot{x}_l - \mathring{x}_l)}{n\cdot s} \cdot \sum_{j=1}^s \sum_{i=1}^n f(\boldsymbol{\dot{x}}(\frac{j}{s})+\boldsymbol{\epsilon}^{(i)}) \frac{\partial \log\pi(\dot{x}_l(\frac{j}{s})+\epsilon^{(i)}_l|x_l)}{\partial x_l}\Big{|}_{x_l=\dot{x}_l(\frac{j}{s})}
    \end{align*}
    The third equation simply rewrites the symbol for a sample from a distribution whose location is described by the explicand as $\boldsymbol{\dot{x}}$ plus some noise $\boldsymbol{\epsilon}$.
    The model outcome in the above formula can be expanded with the Taylor series:
    \begin{equation*}
        f(\boldsymbol{\dot{x}}(\frac{j}{s})+\boldsymbol{\epsilon}^{(i)}) = f(\boldsymbol{\dot{x}}(\frac{j}{s})) + \nabla f(\boldsymbol{\dot{x}}(\frac{j}{s}))^T\cdot \boldsymbol{\epsilon}^{(i)} + O(||\boldsymbol{\epsilon}^{(i)}||^2)
    \end{equation*}
    Substituting the expansion and using $\frac{\partial}{\partial x_l} \log\pi$ as a shorthand for $\frac{\partial \log\pi(\dot{x}_l(\frac{j}{s}) + \epsilon_l|x_l)}{\partial x_l}\Big{|}_{x_l=\dot{x}_l(\frac{j}{s})}$:
    \begin{align*}
        \xi_l &= \frac{(\dot{x}_l - \mathring{x}_l)}{n \cdot s} \sum_{j=1}^s \sum_{i=1}^n [
        f(\boldsymbol{\dot{x}}(\frac{j}{s})) + 
        \nabla f(\boldsymbol{\dot{x}}(\frac{j}{s}))^T\cdot \boldsymbol{\epsilon}^{(i)} + 
        O(||\boldsymbol{\epsilon}^{(i)}||^2)
        ] 
        \frac{\partial}{\partial x_l} \log\pi \\
        &= \frac{(\dot{x}_l - \mathring{x}_l)}{n\cdot s} \cdot
        \Biggl[
            \underbrace{\sum_{j=1}^s \sum_{i=1}^n f(\boldsymbol{\dot{x}}(\frac{j}{s})) \frac{\partial}{\partial x_l} \log\pi }_{\Circled{1}} + 
            \underbrace{\sum_{j=1}^s \sum_{i=1}^n \nabla f(\boldsymbol{\dot{x}}(\frac{j}{s}))^T \cdot \boldsymbol{\epsilon}^{(i)} \cdot \frac{\partial}{\partial x_l} \log\pi }_{\Circled{2}} + 
            \underbrace{\sum_{j=1}^s \sum_{i=1}^n O(||\boldsymbol{\epsilon}^{(i)}||^2) \frac{\partial}{\partial x_l} \log\pi }_{\Circled{3}}
        \Biggr] 
    \end{align*}
    When the number of samples $n$ for the gradient estimator increases, the term \Circled{1} converges in probability to 0 as:
    \begin{align*}
        \Circled{1} &= \sum_{j=1}^s \frac{1}{n} \sum_{i=1}^n f(\boldsymbol{\dot{x}}(\frac{j}{s})) \frac{\partial}{\partial x_l} \log\pi
        = \sum_{j=1}^s \biggl[ f(\boldsymbol{\dot{x}}(\frac{j}{s})) \cdot \frac{1}{n} \sum_{i=1}^n \frac{\partial}{\partial x_l} \log\pi \biggr] \\
        \overset{\mathbb{P}}&{\to} \sum_{j=1}^s \biggl[ f(\boldsymbol{\dot{x}}(\frac{j}{s})) \cdot \mathbb{E}_{\pi(\dot{x}_l(\frac{j}{s}) + \epsilon_l|\dot{x}_l(\frac{j}{s}))}\Bigl[\frac{\partial}{\partial x_l} \log\pi\Bigr] \biggr] = 0 
        % \int_{-\infty}^{+\infty}
    \end{align*}
    This is because the expectation takes a zero value:
    \begin{align*}
        \mathbb{E}_{\pi(\dot{x}_l(\frac{j}{s}) + \epsilon_l|\dot{x}_l(\frac{j}{s}))} \Bigl[\frac{\partial}{\partial x_l} \log\pi \Bigr] &= \int_{-\infty}^{+\infty} \pi(\dot{x}_l(\frac{j}{s})+\epsilon_l|\dot{x}_l(\frac{j}{s})) \cdot \frac{\partial \log\pi(\dot{x}_l(\frac{j}{s}) + \epsilon_l|x_l)}{\partial x_l}\Big{|}_{x_l=\dot{x}_l(\frac{j}{s})} ~\mathrm{d} \epsilon_l \\
        \overset{z_l=\dot{x}_l(\frac{j}{s}) + \epsilon_l}&{=} \int_{-\infty}^{+\infty} \pi(z_l|\dot{x}_l(\frac{j}{s})) \cdot \frac{\partial \log\pi(z_l|x_l)}{\partial x_l}\Big{|}_{x_l=\dot{x}_l(\frac{j}{s})} ~\mathrm{d} z_l \\
        &= \int_{-\infty}^{+\infty} \frac{\partial \pi(z_l|x_l)}{\partial x_l}\Big{|}_{x_l=\dot{x}_l(\frac{j}{s})} ~\mathrm{d} z_l \\ 
        &= \biggl[ \frac{\partial}{\partial x_l} \int_{-\infty}^{+\infty} \pi(z_l|x_l) ~\mathrm{d} z_l \biggr] \bigg{|}_{x_l=\dot{x}_l(\frac{j}{s})} \\
        &= \frac{\partial}{\partial x_l} 1 \Big{|}_{x_l=\dot{x}_l(\frac{j}{s})} \\
        \Rightarrow \mathbb{E}_{\pi(\dot{x}_l(\frac{j}{s}) + \epsilon_l|\dot{x}_l(\frac{j}{s}))} \Bigl[\frac{\partial}{\partial x_l} \log\pi \Bigr] &= 0 \numberthis \label{eq:zero_expect}
    \end{align*}
    The interchange of derivatives and integrals is possible because $\pi(\boldsymbol{z}|\boldsymbol{x})$ is continuously differentiable in $\boldsymbol{x}$, a fundamental prerequisite when choosing the search distribution for gradient estimation.
    Now switching to the term \Circled{2}:
    \begin{align*}
        \Circled{2} &= \sum_{j=1}^s \frac{1}{n} \sum_{i=1}^n \nabla f(\boldsymbol{\dot{x}}(\frac{j}{s}))^T\cdot \boldsymbol{\epsilon}^{(i)} \cdot \frac{\partial}{\partial x_l} \log\pi \\
        &= \sum_{j=1}^s \frac{1}{n} \sum_{i=1}^n \biggl[ \Bigl[ \sum_{k=1}^p \frac{\partial f}{\partial x_{k}} \Big{|}_{\boldsymbol{x} =\boldsymbol{\dot{x}}(\frac{j}{s})} \cdot \epsilon_{k}^{(i)} \Bigr] \frac{\partial}{\partial x_l} \log\pi \biggr] \\
        &= \sum_{j=1}^s \frac{1}{n} \sum_{i=1}^n \biggl[ 
            \frac{\partial f}{\partial x_{l}} \Big{|}_{\boldsymbol{x} =\boldsymbol{\dot{x}}(\frac{j}{s})} \cdot \epsilon_{l}^{(i)}
            + \sum_{\substack{k=1 \\ k\neq l}}^p \frac{\partial f}{\partial x_{k}} \Big{|}_{\boldsymbol{x} =\boldsymbol{\dot{x}}(\frac{j}{s})} \cdot \epsilon_{k}^{(i)}
        \biggr] \cdot \frac{\partial}{\partial x_l} \log\pi \\
        &= \sum_{j=1}^s \underbrace{ \frac{\partial f}{\partial x_{l}} \Big{|}_{\boldsymbol{x} =\boldsymbol{\dot{x}}(\frac{j}{s})} \cdot \frac{1}{n} \sum_{i=1}^n \biggl[ \epsilon_{l}^{(i)} \cdot \frac{\partial}{\partial x_l} \log\pi \biggr]}_{\Circled{\text{a}}}
        + \sum_{j=1}^s \sum_{\substack{k=1 \\ k\neq l}}^p \underbrace{\frac{\partial f}{\partial x_{k}} \Big{|}_{\boldsymbol{x}=\boldsymbol{\dot{x}}(\frac{j}{s})} \cdot \frac{1}{n} \sum_{i=1}^n \biggl[ \epsilon_{k}^{(i)} \cdot \frac{\partial}{\partial x_l} \log\pi \biggr]}_{\Circled{\text{b}}}
    \end{align*}
    The first term \Circled{a} converges to $\frac{\partial f}{\partial x_{l}}\Big{|}_{\boldsymbol{x} =\boldsymbol{\dot{x}}(\frac{j}{s})}$ as observations for the gradient estimator expand:
    \begin{align*}
        \Circled{\text{a}} \overset{\mathbb{P}}&{\to} \frac{\partial f}{\partial x_{l}} \Big{|}_{\boldsymbol{x} =\boldsymbol{\dot{x}}(\frac{j}{s})} \cdot \mathbb{E}_{\pi(\dot{x}_l(\frac{j}{s}) + \epsilon_l|\dot{x}_l(\frac{j}{s}))} \Bigl[ \epsilon_{l} \cdot \frac{\partial}{\partial x_l} \log\pi \Bigr] \\
        &= \frac{\partial f}{\partial x_{l}} \Big{|}_{\boldsymbol{x} =\boldsymbol{\dot{x}}(\frac{j}{s})} 
        \cdot \int_{-\infty}^{+\infty} \epsilon_{l} \cdot \pi(\dot{x}_l(\frac{j}{s})+\epsilon_l|\dot{x}_l(\frac{j}{s})) \cdot \frac{\partial \log\pi(\dot{x}_l(\frac{j}{s}) + \epsilon_l|x_l)}{\partial x_l}\Big{|}_{x_l=\dot{x}_l(\frac{j}{s})} ~\mathrm{d} \epsilon_l \\
        \overset{z_l = \dot{x}_l(\frac{j}{s}) + \epsilon_l}&{=} \frac{\partial f}{\partial x_{l}} \Big{|}_{\boldsymbol{x} =\boldsymbol{\dot{x}}(\frac{j}{s})} 
        \cdot \int_{-\infty}^{+\infty} (z_l - \dot{x}_l(\frac{j}{s})) \cdot \pi(z_l|\dot{x}_l(\frac{j}{s})) \cdot \frac{\partial \log\pi(z_l|x_l)}{\partial x_l}\Big{|}_{x_l=\dot{x}_l(\frac{j}{s})} ~\mathrm{d} z_l \\
        &= \frac{\partial f}{\partial x_{l}} \Big{|}_{\boldsymbol{x} =\boldsymbol{\dot{x}}(\frac{j}{s})} 
        \cdot \int_{-\infty}^{+\infty} (z_l - \dot{x}_l(\frac{j}{s})) \cdot \frac{\partial \pi(z_l|x_l)}{\partial x_l} \Big{|}_{x_l=\dot{x}_l(\frac{j}{s})} ~\mathrm{d} z_l \\
        &= \frac{\partial f}{\partial x_{l}} \Big{|}_{\boldsymbol{x} =\boldsymbol{\dot{x}}(\frac{j}{s})} 
        \cdot \biggl[
        \frac{\partial}{\partial x_l} \int_{-\infty}^{+\infty} (z_l - \dot{x}_l(\frac{j}{s})) \cdot \pi(z_l|x_l) ~\mathrm{d} z_l
        \biggr] \bigg{|}_{x_l=\dot{x}_l(\frac{j}{s})} \\
        &= \frac{\partial f}{\partial x_{l}} \Big{|}_{\boldsymbol{x} =\boldsymbol{\dot{x}}(\frac{j}{s})} \cdot
        \Biggl[
            \biggl[ \frac{\partial}{\partial x_l} \int_{-\infty}^{+\infty} z_l \cdot \pi(z_l|x_l) ~\mathrm{d} z_l \biggr] \bigg{|}_{x_l=\dot{x}_l(\frac{j}{s})}
            - \dot{x}_l(\frac{j}{s}) \cdot \biggl[ \frac{\partial}{\partial x_l} \int_{-\infty}^{+\infty} \pi(z_l|x_l) ~\mathrm{d} z_l \biggr] \bigg{|}_{x_l=\dot{x}_l(\frac{j}{s})}
        \Biggr] \\
        \overset{(\ref{eq:zero_expect})}&{=} \frac{\partial f}{\partial x_{l}} \Big{|}_{\boldsymbol{x} =\boldsymbol{\dot{x}}(\frac{j}{s})} \cdot
        \Biggl[
            \biggl[ \frac{\partial}{\partial x_l} \mathbb{E}_{\pi(z_l|\dot{x}_l(\frac{j}{s}))} \Bigl[z_l\Bigr] \biggr] \bigg{|}_{x_l=\dot{x}_l(\frac{j}{s})} - \dot{x}_l(\frac{j}{s}) \cdot 0
        \Biggr]
        % &= \frac{\partial f}{\partial x_{l}} \Big{|}_{\boldsymbol{x} =\boldsymbol{\dot{x}}(\frac{j}{s})} \cdot \frac{\partial x_l}{\partial x_l} \Big{|}_{x_l=\dot{x}_l(\frac{j}{s})}
    \end{align*}
    Applying the premise stated in Equation~\ref{eq:unbias_distribution} yields:
    % Note, the second term in the expansion represents the linear correlation, therefore any constant bias of the distribution is cancelled.
    \begin{align*}
        \Circled{\mathrm{a}} \overset{\mathbb{P}}&{\to} \frac{\partial f}{\partial x_{l}} \Big{|}_{\boldsymbol{x} =\boldsymbol{\dot{x}}(\frac{j}{s})} \cdot
        \biggl[ \frac{\partial}{\partial x_l} x_l \biggr] \bigg{|}_{x_l=\dot{x}_l(\frac{j}{s})} = \frac{\partial f}{\partial x_{l}} \Big{|}_{\boldsymbol{x} =\boldsymbol{\dot{x}}(\frac{j}{s})}
    \end{align*}
    The second term \Circled{b} produces 0 because of the independent sampling for different features:
    \begin{align*}
        \Circled{\text{b}} \overset{\mathbb{P}}&{\to} \frac{\partial f}{\partial x_{k}} \Big{|}_{\boldsymbol{x}=\boldsymbol{\dot{x}}(\frac{j}{s})} \cdot \mathbb{E}_{\pi(\boldsymbol{\dot{x}}(\frac{j}{s}) + \boldsymbol{\epsilon}|\boldsymbol{\dot{x}}(\frac{j}{s})} \Bigl[ \epsilon_{k} \cdot \frac{\partial}{\partial x_l} \log\pi \Bigr] \\
        \overset{\mathrm{Independency}}&{=} \frac{\partial f}{\partial x_{k}} \Big{|}_{\boldsymbol{x}=\boldsymbol{\dot{x}}(\frac{j}{s})} \cdot \mathbb{E}_{\pi(\dot{x}_k(\frac{j}{s}) + \epsilon_k|\dot{x}_k(\frac{j}{s}))} \Bigl[\epsilon_{k}\Bigr] \cdot \mathbb{E}_{\pi(\dot{x}_l(\frac{j}{s}) + \epsilon_l|\dot{x}_l(\frac{j}{s}))} \Bigl[ \frac{\partial}{\partial x_l} \log\pi \Bigr] \\
        \overset{(\ref{eq:zero_expect})}&{=} \frac{\partial f}{\partial x_{k}} \Big{|}_{\boldsymbol{x}=\boldsymbol{\dot{x}}(\frac{j}{s})} \cdot \mathbb{E}_{\pi(\dot{x}_l(\frac{j}{s}) + \epsilon_l|\dot{x}_l(\frac{j}{s}))} \Bigl[ \epsilon_{k} \Bigr] \cdot 0 \\
        &= 0
    \end{align*}
    Replacing the terms in \Circled{2} with the derived values:
    \begin{equation*}
        \Circled{2} = \sum_{j=1}^s \frac{\partial f}{\partial x_{l}} \Big{|}_{\boldsymbol{x} =\boldsymbol{\dot{x}}(\frac{j}{s})}
    \end{equation*}
    Lastly, the element $O(||\boldsymbol{\epsilon}^{(i)}||^2)$ of the term \Circled{3} is bounded by $c_j^+||\boldsymbol{\epsilon}^{(i)}||^2$ and $c_j^-||\boldsymbol{\epsilon}^{(i)}||^2$, indicating:
    \begin{equation*}
        c_j^-||\boldsymbol{\epsilon}^{(i)}||^2 \leq O(||\boldsymbol{\epsilon}^{(i)}||^2) \leq c_j^+||\boldsymbol{\epsilon}^{(i)}||^2
    \end{equation*}
    where $c_j^-$/$c_j^+$ is a negative/positive constant. 
    The upper and lower bounds for \Circled{3} can then be written as:
    \begin{equation*}
        \sum_{j=1}^s \frac{c_j^-}{n} \sum_{i=1}^n ||\boldsymbol{\epsilon}^{(i)}||^2 \cdot \frac{\partial}{\partial x_l} \log\pi \leq 
        \sum_{j=1}^s \frac{1}{n} \sum_{i=1}^n O(||\boldsymbol{\epsilon}^{(i)}||^2) \cdot \frac{\partial}{\partial x_l} \log\pi  \leq
        \sum_{j=1}^s \frac{c_j^+}{n} \sum_{i=1}^n ||\boldsymbol{\epsilon}^{(i)}||^2 \cdot \frac{\partial}{\partial x_l} \log\pi 
    \end{equation*}
    Expanding the norm in the bound:
    \begin{align*}
        \frac{1}{n} \sum_{i=1}^n ||\boldsymbol{\epsilon}^{(i)}||^2 \frac{\partial}{\partial x_l} \log\pi &= 
        \frac{1}{n} \sum_{i=1}^n  [\epsilon_{l}^{(i)}]^2 \cdot \frac{\partial}{\partial x_l} \log\pi 
        + \sum_{\substack{k=1\\k\neq l}}^p \frac{1}{n} \sum_{i=1}^n [\epsilon_{k}^{(i)}]^2 \cdot \frac{\partial}{\partial x_l} \log\pi \\
        &\overset{\mathbb{P}}{\to} \mathbb{E}_{\pi(\dot{x}_l(\frac{j}{s}) + \epsilon_l|\dot{x}_l(\frac{j}{s}))} \Bigl[ \epsilon_{l}^2 \cdot \frac{\partial}{\partial x_l} \log\pi \Bigr] + \sum_{\substack{k=1\\k\neq l}}^p \mathbb{E}_{\pi(\boldsymbol{\dot{x}}(\frac{j}{s}) + \boldsymbol{\epsilon}|\boldsymbol{\dot{x}}(\frac{j}{s}))} \Bigl[ \epsilon_{k}^2 \cdot \frac{\partial}{\partial x_l} \log\pi \Bigr] \\
        \overset{\mathrm{Independency}}&{=} \mathbb{E}_{\pi(\dot{x}_l(\frac{j}{s}) + \epsilon_l|\dot{x}_l(\frac{j}{s}))} \Bigl[ \epsilon_{l}^2 \cdot \frac{\partial}{\partial x_l} \log\pi \Bigr] + \sum_{\substack{k=1\\k\neq l}}^p \mathbb{E}_{\pi(\dot{x}_k(\frac{j}{s}) + \epsilon_k|\dot{x}_k(\frac{j}{s}))} \Bigl[ \epsilon_{k}^2 \Bigr] \cdot 0 \\
        &= \int_{-\infty}^{+\infty} \epsilon_{l}^2 \cdot \pi(\dot{x}_l(\frac{j}{s}) + \epsilon_l|\dot{x}_l(\frac{j}{s})) \cdot \frac{\partial \log\pi(\dot{x}_l(\frac{j}{s}) + \epsilon_l|x_l)}{\partial x_l}\Big{|}_{x_l=\dot{x}_l(\frac{j}{s})} ~\mathrm{d}\epsilon_l \\
        \overset{z_l = \dot{x}_l(\frac{j}{s}) + \epsilon_l}&{=} \int_{-\infty}^{+\infty} (z_l - \dot{x}_l(\frac{j}{s}))^2 \cdot \pi(z_l|\dot{x}_l(\frac{j}{s})) \cdot \frac{\partial \log\pi(z_l|x_l)}{\partial x_l}\Big{|}_{x_l=\dot{x}_l(\frac{j}{s})} ~\mathrm{d}z_l \\
        &= \int_{-\infty}^{+\infty} (z_l^2 - 2\cdot z_l \cdot \dot{x}_l(\frac{j}{s}) + \dot{x}_l(\frac{j}{s})^2) \cdot \frac{\partial \pi(z_l|x_l)}{\partial x_l} \Big{|}_{x_l=\dot{x}_l(\frac{j}{s})} ~\mathrm{d}z_l \\
        &= \frac{\partial}{\partial x_l}\biggl[
            \int_{-\infty}^{+\infty} z_l^2 \pi(z_l|x_l) ~\mathrm{d}z_l 
            - 2\dot{x}_l(\frac{j}{s}) \int_{-\infty}^{+\infty} z_l \pi(z_l|x_l) ~\mathrm{d}z_l
            + \dot{x}_l(\frac{j}{s})^2 \int_{-\infty}^{+\infty} \pi(z_l|x_l) ~\mathrm{d}z_l 
        \biggr] \bigg{|}_{x_l=\dot{x}_l(\frac{j}{s})} \\
        &= \frac{\partial}{\partial x_l} \biggl[ 
            \mathbb{E}_{\pi(z_l|x_l)}\Bigl[z_l^2\Bigr] 
            - 2\dot{x}_l(\frac{j}{s}) \mathbb{E}_{\pi(z_l|x_l)}\Bigl[z_l\Bigr] 
            + \dot{x}_l(\frac{j}{s})^2 \cdot 1
        \biggr] \bigg{|}_{x_l=\dot{x}_l(\frac{j}{s})} \numberthis\label{eq:the_term_in3}
    \end{align*}
    Denoting the variance of $z_l$ with $\sigma^2_l$, the first term in Equation~\ref{eq:the_term_in3} is:
    \begin{equation*}
        \frac{\partial}{\partial x_l}\mathbb{E}_{\pi(z_l|\dot{x}_l(\frac{j}{s}))} \Bigl[ z_l^2 \Bigr] = \frac{\partial}{\partial x_l}\biggl[ \mathbb{E}_{\pi(z_l|\dot{x}_l(\frac{j}{s}))}^2 \Bigl[ z_l \Bigr] + \sigma_l^2 \biggr] = \frac{\partial}{\partial x_l}\biggl[ (x_l + \delta_l)^2 + \sigma_l^2 \biggr]
    \end{equation*}
    where $\delta$ denotes any bias of the distribution mean $\mathbb{E}_{\pi(z_l|\dot{x}_l(\frac{j}{s}))}[z_l]$ to its location parameter $x_l$. 
    Given that $x_l$, as a location parameter, has no control over the spread $\sigma_l$ of the distribution, Equation~\ref{eq:the_term_in3} then becomes:
    \begin{align*}
        (\ref{eq:the_term_in3}) &= \frac{\partial}{\partial x_l} \biggl[ 
            (x_l + \delta_l)^2 + \sigma_l^2
            - 2\dot{x}_l (x_l + \delta_l)
            + \dot{x}^2
        \biggr] \bigg{|}_{x_l=\dot{x}_l(\frac{j}{s})} \\
        &= \biggl[ 
            2 x_l + 2 \delta_l - 2 \dot{x}_l + 0
        \biggr] \bigg{|}_{x_l=\dot{x}_l(\frac{j}{s})} = 2\cdot\delta_l \\
        \overset{(\ref{eq:unbias_distribution})}&{=} 0
    \end{align*}
    where the unbiasedness of the search distribution ensures the vanishing of the higher order residual.
    % Since the dimension of the input is high (especially for image data), the matrix norm $||\boldsymbol{\epsilon}^{(i)}||$ has a trivial dependency on its \textit{l}-th element:
    % \begin{align*}
    %     \frac{1}{n} \sum_{i=1}^n ||\boldsymbol{\epsilon}^{(i)}||^2 \cdot \frac{\partial}{\partial x_l} \log\pi &\overset{\mathbb{P}}{\to} \mathbb{E}_{\boldsymbol{\dot{x}}(\frac{j}{s})}[||\boldsymbol{\epsilon}^{(i)}||^2 \cdot \frac{\partial}{\partial x_l} \log\pi] \\
    %     &\approx \mathbb{E}_{\boldsymbol{\dot{x}}(\frac{j}{s})} [||\boldsymbol{\epsilon}^{(i)}||^2] 
    %     \cdot \mathbb{E}_{\dot{x}_l(\frac{j}{s})}[\frac{\partial}{\partial x_l} \log\pi] \\
    %     \overset{(\ref{eq:zero_expect})}&{=} \mathbb{E}_{\boldsymbol{\dot{x}}(\frac{j}{s})} [||\boldsymbol{\epsilon}^{(i)}||^2] \cdot 0 = 0
    % \end{align*}
    The upper and lower bounds of \Circled{3} can be updated as:
    \begin{equation*}
        0 \leq \Circled{3} \leq 0 \Rightarrow \Circled{3} = 0
    \end{equation*}
    Combining \Circled{1}, \Circled{2}, and \Circled{3} we show that the gradient estimator converges to the actual gradient without bias, which allows rewriting the explanation as follows:
    \begin{equation} \label{eq:unbiasedness}
        \xi_l = \frac{(\dot{x}_l - \mathring{x}_l)}{s} \cdot \sum_{j=1}^s \frac{\partial f}{\partial x_{l}} \Big{|}_{\boldsymbol{x} =\boldsymbol{\dot{x}}(\frac{j}{s})} 
    \end{equation}
    The aggregation of gradient estimations over the interpolation $\boldsymbol{\dot{x}}(\frac{j}{s})$ approaches the true path integral as the interpolation interval $\frac{1}{s}$ becomes small:
    \begin{align*}
        \xi_l \overset{s\to\infty}&{\to} \int_0^1 \frac{\partial f}{\partial x_l} \frac{\partial x_l}{\partial \alpha} ~\mathrm{d} \alpha
    \end{align*}
    Applying the fundamental theorem for line integrals yields:
    \begin{align*}
        \sum_{k=1}^{p} \xi_k &= \int_0^1 \Bigl(
            \frac{\partial f}{\partial x_1} \frac{\partial x_1}{\partial \alpha}
            + \frac{\partial f}{\partial x_2} \frac{\partial x_2}{\partial \alpha} + ...
            + \frac{\partial f}{\partial x_p} \frac{\partial x_p}{\partial \alpha}
        \Bigr)~\mathrm{d} \alpha \\
        &= \int_0^1 \nabla_{\boldsymbol{x}}f(\boldsymbol{x}(\alpha))~\mathrm{d}\alpha \\
        &= f(\boldsymbol{\dot{x}}) - f(\boldsymbol{\mathring{x}}) \\
        \Rightarrow f(\boldsymbol{\mathring{x}}) + \sum_{k=1}^{p} \xi_k &= f(\boldsymbol{\dot{x}})
    \end{align*}

    % ----------------------------- Another idea for the proof knowing that $\eta$ is an unbiased estimator ---------------------------
    % where $\eta_k$ is an unbiased estimator for the partial derivative with an error of $\sigma$ at $\boldsymbol{x}(\frac{j}{s})$:
    % \begin{align*}
    %     \eta_k(\boldsymbol{x}(\frac{j}{s})) = \frac{\partial}{\partial x_k}J(\boldsymbol{x}(\frac{j}{s})) + \sigma
    % \end{align*}
    % Now rewriting the attribution score by expanding the terms in $\eta_k$:
    % \begin{align*}
    %     \xi_k &= \frac{(x_k - \mathring{x}_k)}{s} \cdot \sum_{j=1}^{s} \frac{\partial}{\partial x_k}J(\boldsymbol{x}(\frac{j}{s})) + \delta_j \\
    %     &= \frac{(x_k - \mathring{x}_k)}{s} \cdot \sum_{j=1}^{s} \frac{\partial}{\partial x_k}J(\boldsymbol{x}(\frac{j}{s})) + \sigma_{\eta} \\
    %     &= \int_0^1 \frac{\partial}{\partial x_k}J(\boldsymbol{x}(\alpha)) \frac{\partial x_k}{\alpha} ~\mathrm{d} \alpha + \sigma_{\xi} + \sigma_{\eta}
    % \end{align*}
    % Denoting the sum of the error terms in $\xi_k$ with $c_k$, applying the fundamental theorem for line integrals yields:
    % \begin{align*}
    %     \sum_{k=1}^{p} \xi_k = \int_0^1 \nabla_{\boldsymbol{x}}J(\boldsymbol{x}(\alpha)) \mathrm{d} \alpha + \sum_{k=1}^{p} c_k
    % \end{align*}
    % Both the errors in $c_k$ converges to 0 as the number of observations increases, meaning that:
    % \begin{align*}
    %     \sum_{k=1}^{p} \xi_k \overset{\mathbb{P}}&{\to} \int_0^1 \frac{\partial}{\partial x_k}J(\boldsymbol{x}(\alpha)) \frac{\partial x_k}{\alpha} \mathrm{d} \alpha = J(\boldsymbol{x}) - J(\boldsymbol{\mathring{x}})        
    % \end{align*}
\end{proof}

In the proof, we show that the total attribution sum of GEEX converges in probability to the prediction difference between the baseline and the explicand.
The error of the explainer arises from two sources: the error of the gradient estimator and the error associated with the approximation of the line integral.
Optimizing the explainer's performance requires minimization of both errors.
% Besides, an extra factor influencing the scale of error is the dimensionality of the input feature space, as the final error $\sum_k c_k$ has a positive linear correlation with the dimensionality $p$.
% This implies that for the explainer to maintain a certain level of precision, in addition to the consideration of the estimation errors, the number of observations requires quadratic growth with a linearly expanding feature space.

\section{Proof of Satisfaction on the Four Axioms} \label{sup:axioms}
Theorem~\ref{theorem:4axioms} states that \methodAbr~fulfills the four fundamental axioms of attribution methods. 
This section gives the proof of these properties one by one.

\subsection{Axiom: Insensitivity}
\textit{Insensitivity} (Dummy) states that the attribution to a feature on which the target model does not functionally depend should be zero.
Formally, for a feature $x_l$, Insensitivity requires:
\begin{equation*}
    \xi_l = 0,~~\text{if}~\frac{\partial f}{\partial x_l}=0~\text{for}~\boldsymbol{x}\in \mathbb{R}^p
\end{equation*}
\begin{proof}
    Focusing on the \textit{l}-th feature that should receive a zero attribution score, its attribution determined by \methodAbr~can be written as follows according to Equation~\ref{eq:unbiasedness}:
    \begin{equation*}
        \xi_l = \frac{(x_l - \mathring{x}_l)}{s} \cdot \sum_{j=1}^s \frac{\partial f}{\partial x_{l}} \Big{|}_{\boldsymbol{x} =\boldsymbol{\dot{x}}(\frac{j}{s})} 
    \end{equation*}
    Then applying the definition of Insensitivity reaches the end of the proof:
    \begin{equation*}
        \xi_l = \frac{(x_l - \mathring{x}_l)}{s} \cdot \sum_{j=1}^s 0 = 0 
    \end{equation*}
\end{proof}

Compared to Insensitivity, a similar but still slightly different property is \textit{Missingness}.
With the absence defined by some baseline value $\mathring{\boldsymbol{x}}$, this property requires attribution methods to distribute a zero value to the contribution $\xi_l$ of an absent feature $x_l$, namely:
\begin{equation*}
    \xi_l = 0,~~\text{if}~x_l=\mathring{x}_l
\end{equation*}
Missingness differs from Insensitvitiy due to its reliance on the definition of a baseline.
However, despite this difference, both properties are carried by \methodAbr.
The proof of Missingness can be done in a single line:
\begin{equation*}
    \xi_l = \frac{(x_l-\mathring{x}_l)}{s}\cdot \sum_{j=1}^{s} \eta_l(\boldsymbol{\dot{x}}(\frac{j}{s})) \overset{x_l=\mathring{x}_l}{=} 0 \cdot \sum_{j=1}^{s} \eta_l(\boldsymbol{\dot{x}}(\frac{j}{s})) = 0 \\
\end{equation*}

\subsection{Axiom: Sensitivity}
\textit{Sensitivity} states that if the explicand and the baseline differing in one feature receive different predictions, then the differing feature should be assigned a non-zero importance score.
The proof of Sensitivity is readily accessible with the help of the proof of Completeness.
% \xi_k &= \frac{(x_k - \mathring{x}_k)}{s} \cdot \sum_{j=1}^{s} \frac{\partial}{\partial x_k}J(\boldsymbol{x}(\frac{j}{s})) + \delta_j \\
        % &= \frac{(x_k - \mathring{x}_k)}{s} \cdot \sum_{j=1}^{s} \frac{\partial}{\partial x_k}J(\boldsymbol{x}(\frac{j}{s})) + \sigma_{\eta}

\begin{proof}
    Denoting the only different feature by $x_{l}$, the explanation outcome takes the following value:
    \begin{equation*}
        \boldsymbol{\xi} = (0, ..., \xi_l, ..., 0)
    \end{equation*}
    because the other terms are canceled out according to Missingness given $x_k = \mathring{x}_k,~\forall x_k\neq x_l$.
    Reusing the conclusion in the proof of Completeness yields:
    % Given that the baseline and the explicand receive different predictions $J(\boldsymbol{x}) \neq J(\boldsymbol{\mathring{x}})$, reusing the derivation in the proof of Completeness yields:
    \begin{align*}
        \xi_l &= \xi_l + \sum^p_{\substack{k=1 \\ k\neq l}} 0 = \sum_{k=1}^{p} \xi_k \\
        \overset{\mathbb{P}}&{\to} \int_0^1 \nabla_{\boldsymbol{x}} f(\boldsymbol{x}(\alpha)) \mathrm{d} \alpha \\
        &= f(\boldsymbol{x}) - f(\boldsymbol{\mathring{x}}) \neq 0 \\
        \Rightarrow \xi_l &\neq 0
    \end{align*}
\end{proof}

\subsection{Axiom: Implementation Invariance}
For any two functionally equivalent models, \textit{Implementation Invariance} indicates that the explanations for the decisions made by the two functionally equivalent models ought to be identical despite the different implementations.
Intuitively, black-box explainers naturally satisfy Implementation Invariance as their explanation procedures do not consider or utilize any details about model implementations.
However, we still give the formal proof for GEEX, which shows that our method aligns with the intuition.

\begin{proof}
    Given two models $f_{\boldsymbol{\phi}_1}(\cdot)$ and $f_{\boldsymbol{\phi}_2}(\cdot)$, functional equivalence indicates:
    \begin{equation*}
        f_{\boldsymbol{\phi}_1}(\boldsymbol{x}) = f_{\boldsymbol{\phi}_2}(\boldsymbol{x}),~\forall \boldsymbol{x}\in\mathbb{R}^p
    \end{equation*}
    where $\boldsymbol{\phi}$ denotes some learnable parameters in a model, which is used to indicate the implementation difference.
    The explanations $\boldsymbol{\xi}^{(f_{\boldsymbol{\phi}_1})}$ and $\boldsymbol{\xi}^{(f_{\boldsymbol{\phi}_2})}$ for the two models at an arbitrary point $\boldsymbol{x}$ hold:
    \begin{align*}
        \boldsymbol{\xi}^{(f_{\boldsymbol{\phi}_1})} &= \frac{(\boldsymbol{x} - \boldsymbol{\mathring{x}})}{n\cdot s} \circ \sum_{j=1}^{s} \sum^{n}_{i=1} f_{\boldsymbol{\phi}_1}(\boldsymbol{z}^{(i)}) \nabla_{\boldsymbol{x}}\log\pi(\boldsymbol{z}^{(i)}|\boldsymbol{x}(\frac{j}{s})) \\
        \overset{f_{\boldsymbol{\phi}_1}(\boldsymbol{x})=f_{\boldsymbol{\phi}_2}(\boldsymbol{x})}&{=} \frac{(\boldsymbol{x} - \boldsymbol{\mathring{x}})}{n\cdot s} \circ \sum_{j=1}^{s} \sum^{n}_{i=1} f_{\boldsymbol{\phi}_2}(\boldsymbol{z}^{(i)}) \nabla_{\boldsymbol{x}}\log\pi(\boldsymbol{z}^{(i)}|\boldsymbol{x}(\frac{j}{s})) \\
        &= \boldsymbol{\xi}^{(f_{\boldsymbol{\phi}_2})}
    \end{align*}
\end{proof}

\subsection{Axiom: Linearity}
For any two functions $f_{\boldsymbol{\phi}_1}(\cdot)$ and $f_{\boldsymbol{\phi}_2}(\cdot)$, \textit{Linearity} requires the explanation for the linear composition of the two functions $af_{\boldsymbol{\phi}_1} + bf_{\boldsymbol{\phi}_2}$ equaling the weighted sum of the separate explanations for them, namely:
\begin{equation*}
    \boldsymbol{\xi}^{(af_{\boldsymbol{\phi}_1}+bf_{\boldsymbol{\phi}_2})} = a\cdot \boldsymbol{\xi}^{(f_{\boldsymbol{\phi}_1})} + b\cdot \boldsymbol{\xi}^{(f_{\boldsymbol{\phi}_2})}
\end{equation*}
The Linearity of \methodAbr~is proved below.
% \textit{Linearity} is fulfilled by \methodAbr~when the loss $\mathcal{L}(f(\cdot))$ is a linear function of model outcomes. For example, \textit{Linearity} holds when \ige~utilizes the raw logits as the loss, i.e. $\mathcal{L}(f(\cdot))=f(\cdot)$, whereas the deployment of cross-entropy loss violates this axiom.
\begin{proof}
    \begin{align*}
        \boldsymbol{\xi}^{(af_{\boldsymbol{\phi}_1}+bf_{\boldsymbol{\phi}_2})} &= \frac{(\boldsymbol{x} - \boldsymbol{\mathring{x}})}{n\cdot s} \circ \sum_{j=1}^{s} \sum^{n}_{i=1} \Big[ 
            a f_{\boldsymbol{\phi}_1}(\boldsymbol{z}^{(i)}) + b f_{\boldsymbol{\phi}_2}(\boldsymbol{z}^{(i)})
        \Bigr] \nabla_{\boldsymbol{x}}\log\pi(\boldsymbol{z}^{(i)}|\boldsymbol{x}(\frac{j}{s})) \\
        &= \frac{a(\boldsymbol{x} - \boldsymbol{\mathring{x}})}{n\cdot s} \circ \sum_{j=1}^{s} \sum^{n}_{i=1} f_{\boldsymbol{\phi}_1}(\boldsymbol{z}^{(i)}) \nabla_{\boldsymbol{x}}\log\pi(\boldsymbol{z}^{(i)}|\boldsymbol{x}(\frac{j}{s})) 
        + \frac{b(\boldsymbol{x} - \boldsymbol{\mathring{x}})}{n\cdot s} \circ \sum_{j=1}^{s} \sum^{n}_{i=1} f_{\boldsymbol{\phi}_2}(\boldsymbol{z}^{(i)}) \nabla_{\boldsymbol{x}}\log\pi(\boldsymbol{z}^{(i)}|\boldsymbol{x}(\frac{j}{s})) \\
        &= a\cdot \boldsymbol{\xi}^{(f_{\boldsymbol{\phi}_1})} + b\cdot \boldsymbol{\xi}^{(f_{\boldsymbol{\phi}_2})} 
    \end{align*}
\end{proof}

\section{Complementary Information from Neighbors on the Path} \label{sup:complement_neighbors}
In the last part of Section~\ref{sec:method}, we finalize the explainer as a dense approximation of the path integral with one-sample gradient estimators:
\begin{equation*}
    \boldsymbol{\xi} = \frac{(\boldsymbol{x} - \boldsymbol{\mathring{x}})}{n^*} \circ \sum_{
    \substack{\boldsymbol{\epsilon}\sim\pi(\cdot|\boldsymbol{0}) \\ \alpha\sim \mathcal{U}_{[0, 1]}}
    } f(\boldsymbol{x}(\alpha) + \boldsymbol{\epsilon}) \nabla_{\boldsymbol{x}}\log\pi(\boldsymbol{\epsilon}|\boldsymbol{0})
\end{equation*}
Intuitively, reducing the capacity of the gradient estimator may seem contradictory to the conclusion drawn from the proof of Completeness, which suggests that optimizing explanation quality requires minimization of estimator errors at both levels.
However, this is not necessarily true because neighboring estimators can share complementary information and thus promote their estimations.
Before showing this complementary information, we first provide a formal definition of ``neighboring estimators''.
The definition relies on the assumption of continuous differentiability of the target function $f(\cdot)$, a fundamental requirement for applying any gradient-based explainers, regardless of the accessibility setting.
\begin{definition}
For an interval $[a, b]$ on a path $\boldsymbol{x}(\alpha)$ in which $f(\cdot)$ is locally linear, the one-sample estimators $\boldsymbol{\eta}_{n=1}(\boldsymbol{x}(\dot{\alpha}))$ for arbitrary points $\dot{\alpha}\in[a,b]$ are \textbf{neighboring} estimators.
\end{definition}
Given the assumption that $f(\cdot)$ is continuously differentiable, it is always possible to find such an interval for any estimator on the path that determines its neighboring estimators.
Next, we prove that the set of neighboring estimators with a size of $m$ achieves the same level of precision as a $m$-sample estimator at some point on the interval $\alpha^*\in[a,b]$  denoted by $\boldsymbol{\eta}_{n=m}(\boldsymbol{x}(\alpha^*))$.
\begin{proof}
    The collaborative estimation of neighbors $\{ \boldsymbol{\eta}(\boldsymbol{x}(\dot{\alpha}^{(i)}) ~|~ \dot{\alpha}^{(i)}\in[a,b] \}$ can be written as:
    \begin{align*}
        \frac{1}{m} \sum_{\dot{\alpha}\sim \mathcal{U}_{[a,b]}} \boldsymbol{\eta}_{n=1}(\boldsymbol{x}(\dot{\alpha}))
        \overset{(\ref{eq:unbiasedness})}&{=} \frac{1}{m} \sum_{\dot{\alpha}\sim \mathcal{U}_{[a,b]}} 
        \biggl[ 
            \nabla_{\boldsymbol{x}} f(\boldsymbol{x}(\dot{\alpha})) + \boldsymbol{\hat{\sigma}}^{(\dot{\alpha})}
        \biggr] \\
        \overset{\text{Local linearity}}&{=} \nabla_{\boldsymbol{x}} f(\boldsymbol{x}(a)) + \frac{1}{m} \cdot \sum_{\dot{\alpha}\sim \mathcal{U}_{[a,b]}} \boldsymbol{\hat{\sigma}}^{(\dot{\alpha})} 
        % \numberthis\label{eq:collaborative_estimate}
    \end{align*}
    where $\boldsymbol{\hat{\sigma}}^{(\dot{\alpha})}$ denotes the estimation error of one estimator.
    As an unbiased estimator, the error term follows some distribution $\mathcal{D}_{(0, \sigma^{(\dot{\alpha})})}$, which has a mean at 0 and a variance of value $[\sigma^{(\dot{\alpha})}]^2$. 
    Recalling that $\boldsymbol{x}$ is the location parameter of the search distribution, the variance of an estimator is:
    \begin{align*}
        [\boldsymbol{\sigma}^{(\dot{\alpha})}]^2 &= \mathbb{V} \Bigl[ f(\boldsymbol{x}(\dot{\alpha}) + \boldsymbol{\epsilon}) \cdot \nabla_{\boldsymbol{x}} \log\pi(\boldsymbol{\epsilon}|\boldsymbol{0})  \Bigr] \\
        \overset{\text{Local linearity}}&{=} \mathbb{V} \Bigl[ 
            (f(\boldsymbol{x}(a) + \boldsymbol{\epsilon}) + \delta(\dot{\alpha}) ) \cdot \nabla_{\boldsymbol{x}} \log\pi(\boldsymbol{\epsilon}|\boldsymbol{0})  
        \Bigr]
    \end{align*}
    where $\delta(\dot{\alpha}) = f(\boldsymbol{x}(\alpha)) - f(\boldsymbol{x}(a))$. Expanding the form above yields:
    \begin{align*}
        [\boldsymbol{\sigma}^{(\dot{\alpha})}]^2 &= \mathbb{V} \Bigl[ 
            f(\boldsymbol{x}(a) + \boldsymbol{\epsilon}) \cdot \nabla_{\boldsymbol{x}} \log\pi(\boldsymbol{\epsilon}|\boldsymbol{0})  
        \Bigr] + \mathbb{V} \Bigl[ 
            \delta(\dot{\alpha}) \nabla_{\boldsymbol{x}} \log\pi(\boldsymbol{\epsilon}|\boldsymbol{0})  
        \Bigr] + 2 (\mathbb{E}\Bigl[ 
            \delta(\dot{\alpha}) f(\boldsymbol{x}(a) + \boldsymbol{\epsilon}) \cdot [\nabla_{\boldsymbol{x}} \log\pi(\boldsymbol{\epsilon}|\boldsymbol{0})]^2
        \Bigr] - 0) \\
        &= \mathbb{V} \Bigl[ 
            f(\boldsymbol{x}(a) + \boldsymbol{\epsilon}) \cdot \nabla_{\boldsymbol{x}} \log\pi(\boldsymbol{\epsilon}|\boldsymbol{0})  
        \Bigr] + \mathbb{E} \Bigl[ 
            [\delta(\dot{\alpha}) \nabla_{\boldsymbol{x}} \log\pi(\boldsymbol{\epsilon}|\boldsymbol{0})]^2
        \Bigr] - 0 + 2 \mathbb{E}\Bigl[ \delta(\dot{\alpha}) f(\boldsymbol{x}(a) + \boldsymbol{\epsilon}) \cdot [\nabla_{\boldsymbol{x}} \log\pi(\boldsymbol{\epsilon}|\boldsymbol{0})]^2 \Bigr] \\
        &= \mathbb{V} \Bigl[ 
            f(\boldsymbol{x}(a) + \boldsymbol{\epsilon}) \cdot \nabla_{\boldsymbol{x}} \log\pi(\boldsymbol{\epsilon}|\boldsymbol{0})  
        \Bigr] + \delta(\dot{\alpha})^2 \mathbb{E} \Bigl[ 
            [\nabla_{\boldsymbol{x}} \log\pi(\boldsymbol{\epsilon}|\boldsymbol{0})]^2
        \Bigr] + 2\delta(\dot{\alpha}) \mathbb{E}\Bigl[ 
            f(\boldsymbol{x}(a) + \boldsymbol{\epsilon}) \cdot [\nabla_{\boldsymbol{x}} \log\pi(\boldsymbol{\epsilon}|\boldsymbol{0})]^2 
        \Bigr]
    \end{align*}
    For a fixed search distribution, the expectations in the last two terms are constant.
    Denoting them as $\boldsymbol{c_1}$ and $\boldsymbol{c_2}$ yields:
    \begin{align*}
        [\boldsymbol{\sigma}^{(\dot{\alpha})}]^2 &= \mathbb{V} \Bigl[ 
            f(\boldsymbol{x}(a) + \boldsymbol{\epsilon}) \cdot \nabla_{\boldsymbol{x}} \log\pi(\boldsymbol{\epsilon}|\boldsymbol{0})  
        \Bigr] + \delta(\dot{\alpha})^2 \boldsymbol{c}_1 + 2\delta(\dot{\alpha}) \boldsymbol{c}_2
    \end{align*}
    where $\boldsymbol{c}_1$ is a constant matrix, and the elements in $\boldsymbol{c}_2$ are not necessarily identical.
    However, mirror sampling adopted in this work ensures the isotropicity of the search distribution, which brings convenience to the proof of complementary information.
    % First, expanding $\boldsymbol{c}_2$ with local linearity gives:
    The isotropic distribution, in combination with local linearity, simplifies $\boldsymbol{c}_2$:
    \begin{align*}
    \boldsymbol{c}_2 &= \mathbb{E}\Bigl[ 
        f(\boldsymbol{x}(a) + \boldsymbol{\epsilon}) \cdot [\nabla_{\boldsymbol{x}} \log\pi(\boldsymbol{\epsilon}|\boldsymbol{0})]^2 
    \Bigr] \\
    \overset{\text{Local linearity}}&= f(\boldsymbol{x}(a)) \cdot \mathbb{E}\Bigl[ 
        [\nabla_{\boldsymbol{x}} \log\pi(\boldsymbol{\epsilon}|\boldsymbol{0})]^2 
    \Bigr] + \mathbb{E}\Bigl[ 
        \nabla f(\boldsymbol{x}(\alpha))^T \cdot \boldsymbol{\epsilon} \cdot [\nabla_{\boldsymbol{x}} \log\pi(\boldsymbol{\epsilon}|\boldsymbol{0})]^2 
    \Bigr] \\
    &= f(\boldsymbol{x}(a)) \cdot \mathbb{E}\Bigl[ 
        [\nabla_{\boldsymbol{x}} \log\pi(\boldsymbol{\epsilon}|\boldsymbol{0})]^2 
    \Bigr] + \sum_{k=1}^p \frac{\partial f}{\partial x_k} \cdot \mathbb{E}\Bigl[ 
        \epsilon_k \cdot [\nabla_{\boldsymbol{x}} \log\pi(\boldsymbol{\epsilon}|\boldsymbol{0})]^2 
    \Bigr] \\
    \overset{\text{Isotropicity}}&{=} f(\boldsymbol{x}(a)) \cdot \mathbb{E}\Bigl[ 
        [\nabla_{\boldsymbol{x}} \log\pi(\boldsymbol{\epsilon}|\boldsymbol{0})]^2 
    \Bigr] + \sum_{k=1}^p \frac{\partial f}{\partial x_k} \cdot \boldsymbol{0}\\
    &= f(\boldsymbol{x}(a)) \boldsymbol{c}_1
    \end{align*}
    The variance of a one-sample estimator can be updated:
    \begin{align*}
        [\boldsymbol{\sigma}^{(\dot{\alpha})}]^2 &= \mathbb{V} \Bigl[ 
            f(\boldsymbol{x}(a) + \boldsymbol{\epsilon}) \cdot \nabla_{\boldsymbol{x}} \log\pi(\boldsymbol{\epsilon}|\boldsymbol{0})  
        \Bigr] + \Bigl[\delta(\dot{\alpha})^2 + 2\delta(\dot{\alpha})\cdot f(\boldsymbol{x}(a)) \Bigr] \boldsymbol{c}_1
    \end{align*}
    The variance of the collaborative estimation denoted by $[\boldsymbol{\sigma}^{[a,b]}_{N=m}]^2$ is:
    \begin{align*}
        [\boldsymbol{\sigma}^{[a,b]}_{N=m}]^2 &= \frac{1}{m^2} \sum_{\dot{\alpha}\sim \mathcal{U}_{[a,b]}} [\boldsymbol{\sigma}^{(\dot{\alpha})}]^2 \\
        &= \frac{1}{m^2} \sum_{\dot{\alpha}\sim \mathcal{U}_{[a,b]}} \mathbb{V} \Bigl[ 
            f(\boldsymbol{x}(a) + \boldsymbol{\epsilon}) \cdot \nabla_{\boldsymbol{x}} \log\pi(\boldsymbol{\epsilon}|\boldsymbol{0})  
        \Bigr] + \Bigl[\delta(\dot{\alpha})^2 + 2\delta(\dot{\alpha})\cdot f(\boldsymbol{x}(a)) \Bigr] \boldsymbol{c}_1 \\
        &= \frac{1}{m} \mathbb{V} \Bigl[ 
            f(\boldsymbol{x}(a) + \boldsymbol{\epsilon}) \cdot \nabla_{\boldsymbol{x}} \log\pi(\boldsymbol{\epsilon}|\boldsymbol{0})  
        \Bigr] + \frac{1}{m^2} \sum_{\dot{\alpha}\sim \mathcal{U}_{[a,b]}} \Bigl[\delta(\dot{\alpha})^2 + 2\delta(\dot{\alpha})\cdot f(\boldsymbol{x}(a)) \Bigr] \boldsymbol{c}_1 \\
        &= \frac{1}{m} \mathbb{V} \Bigl[ 
            f(\boldsymbol{x}(a) + \boldsymbol{\epsilon}) \cdot \nabla_{\boldsymbol{x}} \log\pi(\boldsymbol{\epsilon}|\boldsymbol{0})  
        \Bigr] + \frac{1}{m^2} \sum_{\dot{\alpha}\sim \mathcal{U}_{[a,b]}} \Bigl[
            f(\boldsymbol{x}(\alpha))^2 - f(\boldsymbol{x}(a))^2
        \Bigr] \boldsymbol{c}_1 
    \end{align*}
    Similarly, the variance of estimation error for an $m$-sample estimator is:
    \begin{align*}
        [\boldsymbol{\sigma}^{(\alpha)}_{N=m}]^2 &= \frac{1}{m} \mathbb{V} \Bigl[ 
            f(\boldsymbol{x}(a) + \boldsymbol{\epsilon}) \cdot \nabla_{\boldsymbol{x}} \log\pi(\boldsymbol{\epsilon}|\boldsymbol{0})  
        \Bigr] + \frac{1}{m} \Bigl[
            f(\boldsymbol{x}(\alpha))^2 - f(\boldsymbol{x}(a))^2
        \Bigr] \boldsymbol{c}_1 
    \end{align*}
    Depending on the value of $f(\cdot)$ on the interval, the standard deviation of the collaborative estimator is bounded either by both endpoints of the segment when the minimum of $f(\boldsymbol{x}(\alpha))^2 - f(\boldsymbol{x}(a))^2$ is not on $[a, b]$:
    \begin{equation*}
        \min (\boldsymbol{\sigma}^{(a)}_{N=m}, \boldsymbol{\sigma}^{(b)}_{N=m}) \leq \boldsymbol{\sigma}^{[a,b]}_{N=m} \leq \max (\boldsymbol{\sigma}^{(a)}_{N=m}, \boldsymbol{\sigma}^{(b)}_{N=m})
    \end{equation*}
    or by the minimum $\alpha_{min}\in[a,b]$ and the endpoint with a higher function outcome otherwise:
    \begin{equation*}
        \boldsymbol{\sigma}^{(\alpha_{min})}_{N=m} \leq \boldsymbol{\sigma}^{[a,b]}_{N=m} \leq \max (\boldsymbol{\sigma}^{(a)}_{N=m}, \boldsymbol{\sigma}^{(b)}_{N=m})
    \end{equation*}
    Given the continuous differentiability and the bounded variance on the interval, there always exists a point $\alpha^*\in[a,b]$ such that:
    \begin{align*}
        &\Bigl[ f(\boldsymbol{x}(\alpha^*))^2 - f(\boldsymbol{x}(a))^2 \Bigr] \boldsymbol{c}_1 
        = \frac{1}{m} \sum_{\dot{\alpha}\sim \mathcal{U}_{[a,b]}} \Bigl[
            f(\boldsymbol{x}(\alpha))^2 - f(\boldsymbol{x}(a))^2
        \Bigr] \boldsymbol{c}_1 \\
        \Rightarrow & ~[\boldsymbol{\sigma}^{(\alpha^*)}_{N=m}]^2 = [\boldsymbol{\sigma}^{[a,b]}_{N=m}]^2
    \end{align*}
    In other words, the collaborative estimator achieves the same level of precision as the $m$-sample estimator.
\end{proof}

\end{document}